\documentclass{article} % For LaTeX2e
\usepackage[numbers]{natbib}

\usepackage[preprint]{nips_2018}

\usepackage{amsmath,amsfonts,bm}

\def\eqref#1{equation~\ref{#1}}
\def\1{\bm{1}}

\def\ve{{\bm{e}}}

\def\vl{{\bm{l}}}

\def\vr{{\bm{r}}}

\DeclareMathAlphabet{\mathsfit}{\encodingdefault}{\sfdefault}{m}{sl}
\SetMathAlphabet{\mathsfit}{bold}{\encodingdefault}{\sfdefault}{bx}{n}

\usepackage{amssymb}
\usepackage{hyperref,url,subcaption}
\usepackage[ruled,vlined]{algorithm2e}
\usepackage{booktabs,multirow,wrapfig}
\usepackage[font={small}]{caption}

\usepackage[colorinlistoftodos,textsize=tiny]{todonotes}

\usepackage{titlesec}
\titlespacing{\paragraph}{%
  0pt}{%              left margin
  0.3em}{% space before (vertical)
  1em}%               space after (horizontal)

\usepackage{lipsum}
\newcommand\blfootnote[1]{%
  \begingroup
  \renewcommand\thefootnote{}\footnote{#1}%
  \addtocounter{footnote}{-1}%
  \endgroup
}

\title{Reinforced Imitation in Heterogeneous Action Space}

\author{
Konrad \.Zo\l{}na$^{1,2,\dagger}$ {} {} {} {} {} {} Negar Rostamzadeh$^{2}$ {} {} {} {} {} {} Yoshua Bengio$^{3,4}$\\
{\bf Sungjin Ahn$^{5,\ddagger}$ {} {} {} {} {} {} Pedro O. Pinheiro$^{2,\ddagger}$}\\\\
$^1$Jagiellonian University {} {} {} {} {} {} $^2$Element AI {} {} {} {} {} {} $^3$Universit\'e de Montr\'eal, MILA \\
$^4$CIFAR Senior Fellow {} {} {} {} {} {} $^5$Rutgers University
}

\begin{document}
\maketitle

\blfootnote{$^{\dagger}$Corresponding author: \texttt{konrad.zolna@gmail.com}. $^{\ddagger}$Equal advising.}

%%%%%%%%%%%%%%%%%%%%%%%%%%%%%%%%%%%%%%%%%%%%%%%%%%%%%%%%%%%%%%%%%%%%%%%%%%%%%%%%%%%%%%%%%%%%%%
% Abstract
%%%%%%%%%%%%%%%%%%%%%%%%%%%%%%%%%%%%%%%%%%%%%%%%%%%%%%%%%%%%%%%%%%%%%%%%%%%%%%%%%%%%%%%%%%%%%%
\begin{abstract}
Imitation learning is an effective alternative approach to learn a policy when the reward function is sparse. In this paper, we consider a challenging setting where an agent and an expert use different actions from each other. We assume that the agent has access to a sparse reward function and state-only expert observations. We propose a method which gradually balances between the imitation learning cost and the reinforcement learning objective. In addition, this method adapts the agent's policy based on either mimicking expert behavior or maximizing sparse reward.
We show, through navigation scenarios, that (i) an agent is able to efficiently leverage sparse rewards to outperform standard state-only imitation learning, (ii) it can learn a policy even when its actions are different from the expert, and (iii) the performance of the agent is not bounded by that of the expert, due to the optimized usage of sparse rewards.
\end{abstract}

%%%%%%%%%%%%%%%%%%%%%%%%%%%%%%%%%%%%%%%%%%%%%%%%%%%%%%%%%%%%%%%%%%%%%%%%%%%%%%%%%%%%%%%%%%%%%%
% Introduction
%%%%%%%%%%%%%%%%%%%%%%%%%%%%%%%%%%%%%%%%%%%%%%%%%%%%%%%%%%%%%%%%%%%%%%%%%%%%%%%%%%%%%%%%%%%%%%
\section{Introduction}\label{sec:introduction}
Learning by imitation is one of the most fundamental forms of learning in nature~\citep{jones2009development, david1999animal}. Its critical role in cognitive development is also supported by the fact that human brains have special structures, such as mirror neurons, which are presumed to support this ability~\citep{heyes2010mirror}. Due to this significance, it has also played a key role in machine learning and robotics~\citep{pomerleau89alvinn,ratliffBS07}, especially for problems in which reinforcement learning (RL) can easily become inefficient, \emph{e.g.}, as in the case of sparse reward signals. 

Imagine a toddler (the \emph{learner}) observing a caregiver (the \emph{expert}) who is performing a task, \emph{e.g.}, opening a door. From this example, we can derive the following observations on Imitation Learning (IL). First, unlike the typical imitation learning in machine learning, the true action labels (motor command) executed by the expert is not directly provided to the learner. Although in some cases such as autonomous cars, it may be possible to have access to the internal action labels by deploying special equipment, it is still expensive in general and in many applications not possible at all. 
Second, the actions that can be executed by the expert and the learner are different because of the differences in body development. This challenge can also occur in real world applications. For instance, there may be a new version of a home robot that needs to learn from demonstrations of the old version which supports only a naive set of actions, \emph{e.g.}, only a subset of the new version. Third, it may be reasonable and realistic to augment imitation learning with sparse reward signals. Even if having access to labels for every action is unrealistic, as in many sparse rewards cases, \emph{e.g.}, the completion of a task signaled by language or facial expression, can easily and cheaply be obtained. These challenges together with the inherent difficulties of reinforcement learning such as the sparsity of the reward signal and sensitivity to hyperparameter tuning, are required to be dealt with jointly, in order to make imitation learning applicable to real and complex challenges.

In this paper, we propose a method for \emph{Reinforced Imitation Learning from Observations} (RILO) to tackle the aforementioned challenges. Following the above observations, the proposed method aims to work efficiently in a setting where (i) the expert and learner operate in different action spaces, (ii) labels for expert actions are not available, and (iii) a reward signal is only sparsely provided. To achieve this, we extend generative adversarial imitation learning (GAIL)~\citep{ho2016generative} to improve efficiency in cases where the expert actions are not available \textit{and} different from the learner actions. The proposed approach can automatically balance learning between imitation and environment rewards. Our model gradually, but automatically, releases the reliance on the imitation reward and then learns more from the environment rewards. Additionally, the learner is able to alter the policy depending on the objective of mimicking expert behavior or maximizing sparse rewards. These two features make our method learn as fast as imitation learning but also potentially converge to a better policy than that of the expert demonstration.

Through a series of experiments in navigation scenarios (both fully and partially observable), also in high dimensional pixel space, we show that an agent is able to combine both the state observations from an expert and the sparse rewards to achieve better performance than either pure RL or IL with state-only observations. The main contribution of the paper is as follows.
First, we provide an algorithm, dubbed \emph{self-exploration}, that can be applied (and is suitable) when expert and learner do not share the same action space and state-only observations are provided. Second, we validate its effectiveness through a series of systematic experiments, analyzing and comparing to previously proposed methods. Finally, we show that the performance of the learner is not bounded by that of the expert due to suitable use of the environment sparse rewards.

%%%%%%%%%%%%%%%%%%%%%%%%%%%%%%%%%%%%%%%%%%%%%%%%%%%%%%%%%%%%%%%%%%%%%%%%%%%%%%%%%%%%%%%%%%%%%%
% Related Works
%%%%%%%%%%%%%%%%%%%%%%%%%%%%%%%%%%%%%%%%%%%%%%%%%%%%%%%%%%%%%%%%%%%%%%%%%%%%%%%%%%%%%%%%%%%%%%
\section{Related Work}\label{sec:related-work}
Imitation learning (IL) is a common approach to learn a policy from expert demonstrations, \emph{i.e.} sequences of state-action pairs. IL includes two main categories: (i) behavioural cloning~\citep{bainS95,pomerleau89alvinn} and (ii) inverse reinforcement learning~\citep{abbeel2004appren,ng2000ilr}.~Behavioral cloning directly learns the mapping from a state to an action by using the true action-labels from demonstrations and thus represents a supervised learning method. Inverse reinforcement learning derives a reward function from demonstrations,  that can then be used to train a policy using the learned reward function.

Recently,~\citet{ho2016generative} proposed GAIL, a method that uses demonstration data by forcing the learner to match state-transition occupancy distribution of the expert using an approach similar to GANs~\citep{goodfellow2014gan}. Although GAIL is very effective and many work have been built on it \citep{li2017infogail,stadie2017third}, this method requires expert state-action pairs, which are expensive to obtain in many applications. 

For this reason, we focus on an IL scenario in which an agent does not use the expensive true action labels from an expert, but instead only uses state observations. Following the previous literature, we call this \textit{imitation learning from observation} (ILO). While there are only a few existing studies focusing  on this problem~\citep{aytar2018playing,kimura2018internal,liu2017imitation,stadie2017third,torabi2018behavioral}, there are two that are closest to our method \citep{merel2017learning,torabi2018generative}. While both methods are built on top of GAIL, unlike GAIL, they target the case where state-only observations are provided. Contrary to our approach, these methods work under the pure IL setting, \emph{i.e.} they do not take advantage of the potential availability of sparse rewards.
Some other line of works consider state-only expert observations during training, but also require expert observations at test time~\citep{borsa2017observational,duan2017one,pathak2018zero}.

Demonstrations can also be used to improve the exploration efficiency of RL under sparse reward settings~\citep{hester2017deep,nair2017overcoming,vecerik2017leveraging}. Unlike our approach, these studies use  expensive state-action paired demonstrations, and treat them as self-generated. 
These methods also consider the optimal demonstrations only. While \citet{kang2018policy} and \citet{zhu2018reinforcement} do not rely on these assumptions, they still use state-action demonstrations and assume that both expert and learners share the same action space. They also used the full reward by combining the environment reward and imitation reward as a convex combination whose weights are manually set as a hyperparameter. Our method can, however, automatically adapt the balance and the learner's policy based on either mimicking expert behavior or maximizing sparse reward.

The last two works that we would like to mention are \citep{gao2018reinforcement,gupta2017learning}.
The former considers agents that may be morphologically distinct, however, their approach assumes that time alignment is trivial which does not hold in our experimental setup. The latter works on imperfect demonstrations which makes the work related to our (considering different actions spaces implies dealing with imperfect demonstrations), however their approach assumes the access to the demonstrations (including expert actions).

%%%%%%%%%%%%%%%%%%%%%%%%%%%%%%%%%%%%%%%%%%%%%%%%%%%%%%%%%%%%%%%%%%%%%%%%%%%%%%%%%%%%%%%%%%%%%%
% Method
%%%%%%%%%%%%%%%%%%%%%%%%%%%%%%%%%%%%%%%%%%%%%%%%%%%%%%%%%%%%%%%%%%%%%%%%%%%%%%%%%%%%%%%%%%%%%%
\section{Method}\label{sec:method}
Our method is based on the recently proposed GAIL~\citep{ho2016generative}.
This method employs an adversarial training approach as a way to make the distribution of state-action pairs of the learner as indistinguishable as possible from that of the expert. 
In contrast to GAIL, we do not make the following strong assumption: (i) the learner has access to the actions of the expert and (ii) the expert and the learner possess the same action spaces.

% Differences between our setting and standard GAIL
Our setting differs from the standard GAIL approach in three ways. First, we assume that state-only expert observations are provided as a dataset of trajectories $T_e=\{(s^i_1,...,s^i_{n_i})\}_{i=1}^N$. These trajectories consist of observations derived from executing an expert policy on a (possibly partially observable) Markov decision process with state space $\mathcal{S}$, action space $\mathcal{A}$, a transition probability $p(s_{t+1}|s_t,a_t)$, a reward function $r$, and discount factor $\gamma$. In our setting, an expert policy $\pi_e(a_t|s_t)$ performs actions in $\mathcal{A}^e \subset \mathcal{A}$.
Second, we also consider the learner to have actions in $\mathcal{A}^l\subset\mathcal{A}$, potentially different from $\mathcal{A}^e$. Note that both the expert and the learner operate on the same state space $\mathcal{S}$ and we assume the same transition probability as defined by the superset of agents' action spaces, \emph{i.e} $\mathcal{A}^e \cup \mathcal{A}^l \subseteq \mathcal{A}$.
Finally, since the expert and the learner can perform different actions, the two agents may have different optimal policies. In this case, pure imitation learning methods would end up with a learner having a sub-optimal policy. We propose \emph{self-exploration} method to utilize the availability of sparse environment rewards to escape from the sub-optimal policy (see Subsection~\ref{subsec:se}).

% Overview of method (then go into details later)
\subsection{Overview}\label{overview}
\begin{figure}[t]
\begin{center}
\centerline{\includegraphics[width=.7\linewidth]{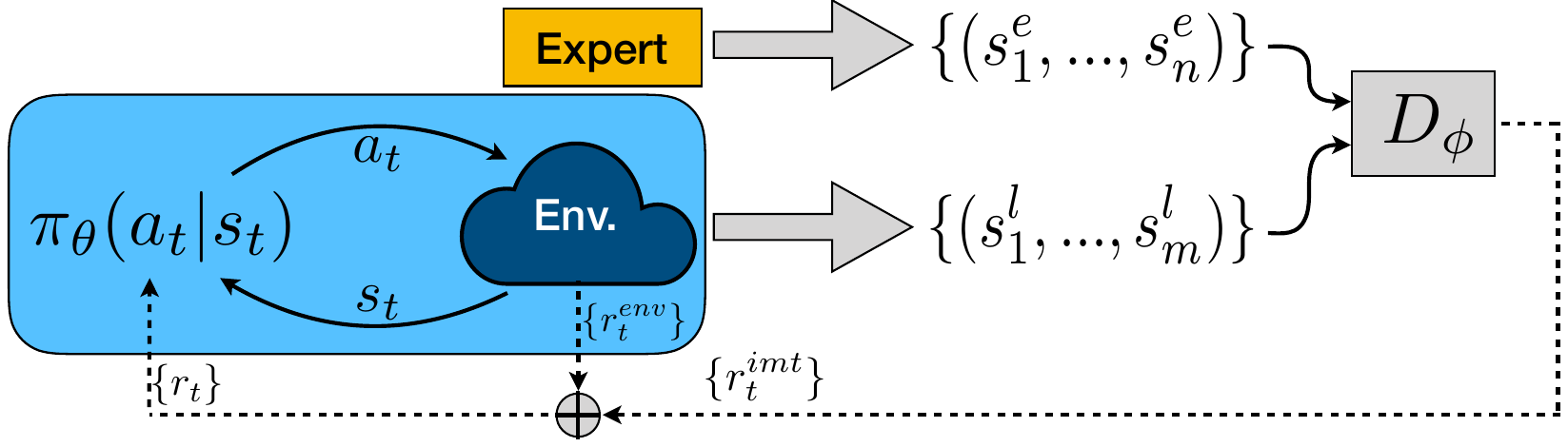}}
\caption{A representation of RILO framework. An agent (learner) with a policy $\pi_\theta$ interacts with an environment producing a trajectory of states and a sparse reward. Imitation rewards are acquired by comparing state-only observations from the learner and the expert. The policy is updated with a combination of both rewards.}
\label{fig:schematic}
\end{center}
\vskip -0.15in
\end{figure}

Our approach is composed of two different components (see Figure~\ref{fig:schematic}). A \textit{policy} $\pi_{\theta}:\mathcal{S}\to\mathcal{A}^l$ outputs an action given its current state. A \textit{discriminator} $D_{\phi}:\mathcal{S}\times\mathcal{S}\to[0,1]$ is responsible for identifying the agent (either expert or learner) that has generated a given pair of states.%from a trajectory comes from.

Our goal is to optimize the following minimax objective:
\begin{equation*}
\min_{\theta} \; \max_{\phi}\; \mathbb{E}_{\pi_\theta}[\log(1 - D_\phi(s_i,s_j))] + \mathbb{E}_{T_e}[\log(D_\phi(s_i,s_j))] \;,
\end{equation*}
where $(s_i,s_j)$ is a tuple of states generated by the policy $\pi_\theta$ or $\pi_e$. Similar to~\cite{zhu2018reinforcement}, the policy is trained to maximize the discounted sum of the final reward $r_t$, which is the sum of the environment reward $r^{env}_t$ and the imitation reward $r^{imt}_t$ given by the discriminator:
\begin{equation}\label{final_reward}
    r_t = r^{env}_t + \lambda \cdot r^{imt}_t\;,\;\; \lambda>0 \;.
\end{equation}

In the next subsection, we describe how we define the imitation reward $r^{imt}_t$. The reward $r_t$ from Equation~\ref{final_reward} encourages the learned policy to visit similar paths as the expert, while obtaining high environment reward by achieving the goal. In all our experiment we always used $\lambda=1$.

% General steps on how to do the update
Each policy iteration consists of the following steps: (i) collect observations with the current learner policy, (ii) compute the discriminator scores for pairs of states where each pair is from either the expert or the learner, and update the discriminator, and (iii) compute the final rewards $\{r_t\}$ and update the policy.
See Figure~\ref{fig:schematic} for the illustration of this procedure.

For the rest of this section, we describe the main contributions that we found to be fundamental to make an agent learn to complete a task in the RILO setting.

%%%%%%%%%%%%%%%%%%%%%%%%%%%%%%
% Imitation Rewards
%%%%%%%%%%%%%%%%%%%%%%%%%%%%%%
\subsection{Imitation Rewards}
\begin{table*}[t]
\caption{Comparison of different imitation rewards. Each method assumes a different input for the discriminator. GAIL considers action-state pairs and is shown only for reference. The following methods assume, respectively, consecutive pairs of states, a single state only, a pair with a given state and random previous state, and all possible pairs of states containing a given state. Each method is responsible for an imitation reward. In RTGD, $\rho(t)$ returns a random integer smaller than $t$. In ATD, $D^*$ is a clamped version of $D$ that never returns a value smaller than the average of all states. \textit{Normalize} operation is described in Subsection \ref{subsec:normalization}.}
\label{tab:dis_rewards}
\begin{center}
\begin{small}
\begin{sc}
\begin{tabular}{l||c|c|c|c|c}
& \textbf{GAIL} & \textbf{CSD} & \textbf{SSD} & \textbf{RTGD} & \textbf{ATD} \\
\midrule
\textbf{Input} & $(a_t, s_t)$ & $(s_{t-1}, s_t)$ & $(s_t,0)$ & $(s_{\rho(t)},s_t)$ & $\{( s_i, s_t) | i \neq t\}$\\[0.15cm]
\textbf{Score} $\mu_t$ & $1 - D(a_t, s_t)$ & $1 - D(s_{t-1}, s_t)$ & $1 - D(s_t)$ & $1 - D(s_{\rho(t)}, s_t)$ & $ \frac{\sum_{i \neq t} (1 - D^*(s_i, s_t))}{(T-1)}$\\[0.15cm]
\midrule
\textbf{Reward} $r^{imt}_t$ & \multicolumn{5}{c}{$Normalize(- \log \mu_t$)}\\
\end{tabular}
\end{sc}
\end{small}
\end{center}
\vskip -0.1in
\end{table*}

As mentioned above, contrary to GAIL, we do not take into account state-action pairs but focus solely on state observations.~\citet{torabi2018generative} use pairs of consecutive transition states as a proxy to encode unobserved actions from expert. As we show in the experiment section, this strategy fails when two agents have different action spaces. In this case, the discriminator can easily discriminate between the expert and the learner, because short-term state transitions provide strong information about the agents' actions. We call this approach Consecutive States Discrimination (CSD).

Other approaches~\citep{merel2017learning,zhu2018reinforcement} provide only the current state, by simply ignoring the action $a_t$ used in the original GAIL approach. The main limitation of this method is that the discriminator is not aware of the dynamics of the agent's trajectory. As we show later, this approach requires a large amount of expert observations to provide a reasonable performance. We dub this method Single State Discrimination (SSD).

We consider CSD and SSD as our baseline methods and analyze two alternatives for the input to the discriminator that circumvent the aforementioned issues.

We call the next method Random Time Gap Discrimination (RTGD). In RTGD, a pair of states is chosen with random time gaps. 
Like CSD, this simple and effective method retains information about the agent's trajectory dynamics. Unlike CSD, it avoids its limitations to very short-term transitions. resulting in state-pairs that are not trivial to the discriminator.

Another solution would be to consider all state pairs instead of a single one. In this case, the final reward at time $t$ is based on the discriminator scores of all pairs containing state $s_t$. However, that would still give the discriminator many short-term transitions. A naive way would be to exclude them using a threshold parameter, as done in the RTGD. 

Average per Time Discrimination (ATD), as we call the fourth method analyzed in this work, makes a better use of these scores to improve discrimination. First, this method compute the mean of the scores of all pairs. Then, the lowest discrimination scores (that is, the pairs in which the discriminator is more confident that it comes from the learner) are clamped with the mean value. As a result, ATD does not rely on any hyperparameters.
We assume that the pairs that are easily identified by the discriminator are scored low due to the different action spaces, rather than a bad long-term strategy\footnote{We indeed observe this in our experiment. Short-term transitions are much more likely to obtain a low score when action spaces for the learner and the expert differs.}. 

The imitation reward $r^{imt}_t$, which measures the similarity between the learner and the expert policies, depends on the method used for constructing the input. 
The full comparison of the imitation rewards is shown in Table~\ref{tab:dis_rewards}.

%%%%%%%%%%%%%%%%%%%%%%%%%%%%%%
% Efficient Combination of RL and IL
%%%%%%%%%%%%%%%%%%%%%%%%%%%%%%
\subsection{Self-Exploration}\label{subsec:se}
Since action spaces between two agents are not necessarily similar, it is unlikely that an optimal policy for the learner is the same as that for the expert. For example, imagine a situation in which an expert's action space is a subset of the learner's action space, $\mathcal{A}^e \subset\mathcal{A}^l$ (\emph{e.g.}, grid world in which the learner can move to all eight adjacent directions while the expert can only move on the four perpendicular directions). In this case, the learner can be penalized by the discriminator for performing %the more efficient
actions in $\mathcal{A}^l \setminus \mathcal{A}^e$ (diagonal moves in the example) because it can easily be distinguished by the discriminator, even though those actions are optimal.

To resolve this issue, we propose to give the learner the possibility to explore the environment by being free from imitating the expert's behaviour. As a result, the final form of Equation \ref{final_reward} becomes:
\begin{equation*}
    r_t = r^{env}_t + \lambda\cdot \tau_k\cdot r^{imt}_t\;, \hspace{1mm} \text{and}\hspace{1mm}\tau_k \sim \text{Bernoulli}(1-\upsilon_{k-1}),
\end{equation*}
where $\upsilon_{k-1}$ is estimated success rate. That is, the self-exploration parameter $\tau_k$ is a binary random variable controlling whether to consider $r_t^{imt}$ or not.  Importantly, the learner is aware of the value $\tau_k$, when it performs its actions which is realized by adding the binary feature to the input. It means that the agent can learn to behave differently when asked to imitate the expert or when it is not under supervision (and maximize environment rewards only).

\begin{wrapfigure}{r}{0.59\textwidth}
\vspace{-0.6cm}
\begin{algorithm}[H]
\SetAlCapFnt{\small}
\SetAlCapNameFnt{\small}
\small

\SetKwInOut{Input}{input}
\SetKwInOut{Output}{output}
\Input{Set of expert trajectories $T_e$, the coefficient $\lambda > 0$,\\
initial policy and discriminator parameters $\theta_{0}$ and $\phi_{0}$.}
\Output{Policy $\pi_{\theta_K}$.}
Initialize a success rate estimate $\upsilon_0$ to be $0$.

\For{$k \leftarrow 1$ \KwTo $K$}{
Sample $\tau_k \sim Bernoulli(1 -\upsilon_{k-1})$.

Get observations $\vl = (s^l_0, ..., s^l_m) \sim \pi_{\theta{k-1}}(\tau_k)$ and environment rewards $\vr^{env} = (r^{env}_1,..., r^{env}_m)$.

Update success rate estimate $\upsilon_k$.

\If{$\tau_k = 1$}{
    Sample expert trajectory $\ve = (s^e_0, s^e_1, ..., s^e_n) \sim T_{e}$.
    
    Compute discriminator scores for expert and learner pairs of states:\\
   $\qquad\mathcal{D}_e = \{d^e_{i,j} = D_{\phi_{k-1}}(s^e_i, s^e_j): i \neq j\}\;\;$ and\\
   $\qquad\mathcal{D}_l = \{d^l_{i,j} = D_{\phi_{k-1}}(s^l_i, s^l_j): i \neq j\}$.
    
    Update  $D_{\phi_{k}}$  to minimize $BCE(\mathcal{D}_e, 1) + BCE (\mathcal{D}_l, 0)$.
    
    Use $\mathcal{D}_l$ to build imitation rewards (normalized):\\
   $\qquad\vr^{imt} = (r^{imt}_1, ..., r^{imt}_m)$.
    
    Construct the final rewards $\vr = \vr^{enc} + \lambda\vr^{imt}$.
}\Else{
    Use environment rewards as the final rewards $\vr = \vr^{env}$.
}
Update  $\pi_{\theta_{k}}$ with the final rewards $\vr$ and any RL-algorithm.
}
\caption{RILO training procedure}
\label{alg}
\end{algorithm}
\vspace{-1.2cm}
\end{wrapfigure}

We set $\upsilon_k$ to be the estimate of the current learner policy success rate\footnote{In our experiments, we used a moving average to compute the success rate $\upsilon_k$.} and thus make the imitation reward guide the learning in the early stage. As the policy gains maturity, \emph{i.e.} as the behaviour of the learner approaches that of the expert trajectory, the success rate will increase accordingly and the policy will then learn more from environment rewards. In other words, we want our agent to become independent of and not limited by the expert's supervision that may, as argued before, become harmful at some point during training. We empirically show that allowing the learner to interact with the environment without imitating, leads to better results and the learner is more likely to use actions from $\mathcal{A}^l \setminus \mathcal{A}^e$.

Algorithm~\ref{alg} shows pseudocode for the details of the training procedure.

\subsection{Normalized Imitation Rewards} \label{subsec:normalization}
To the best of our knowledge, prior work that combine GAIL-based imitation rewards and environment rewards \cite{torabi2018generative,zhu2018reinforcement} use unnormalized imitation rewards, \emph{i.e.} $r_t^{imt} = - \log \mu_t$, where $\mu_t$ is the discriminator score. However, this reward is always positive (since $\mu_t \in (0, 1)$). This may have a negative influence on the agent's behaviour, \emph{e.g.} an agent may tend to walk around the goal but never reach it. We checked empirically that the use of unnormalized imitation rewards leads to very unstable training procedure that usually fails.

Hence, we argue that imitation rewards should be normalized (especially when combined with environment rewards). 
In our experiments, reward is normalized in two different ways, depending on the task complexity: (i) by using a fixed value of $- \log 0.5$ (assuming that $0.5$ is the average discriminator score) or (ii) by subtracting the dynamic value computed as the average of unnormalized imitation rewards for each batch (hence adding imitation rewards does not change the sum of the final rewards).

%%%%%%%%%%%%%%%%%%%%%%%%%%%%%%%%%%%%%%%%%%%%%%%%%%%%%%%%%%%%%%%%%%%%%%%%%%%%%%%
% Experiments
%%%%%%%%%%%%%%%%%%%%%%%%%%%%%%%%%%%%%%%%%%%%%%%%%%%%%%%%%%%%%%%%%%%%%%%%%%%%%%%
\section{Experiments}\label{sec:experimental}
In this section, we show that our approach performs favorably in the RILO setting.
Our experiments aim to show that the agent can combine sparse, unshaped environment rewards with state-only observations (from the expert) to succeed in navigation tasks. The specific questions we address are as follow.
(i) Is self-exploration effective?
(ii) How does the performance of the analyzed methods change when the action space of the expert and the learner are different?
(iii) Can the learner improve over the expert?

We conduct our main experiments in two grid world environments: a fully observable grid world (we call it \emph{F grid world}) and a partially observable one (\emph{P grid world}). 
Using grid world environments allows us to perform rich, systematic set of experiments.
Additionally, to demonstrate that our results generalize to 3D environments, we also experiment on a more complex environment using ViZDoom engine \citep{KempkaWRTJ16} (see Subsection \ref{subsec:vizdoom}).

All environments are designed in a way that a standard RL agent can succeed tasks when the reward is dense and always fails when the reward is sparse.

%Due to limited space, we defer implementation details to Appendix~\ref{app:implementation}. Source code is available in the supplementary material\footnote{Source code will be released upon acceptance.}.
We defer implementation details to Appendix~\ref{app:implementation}

%%%%%%%%%%%%%%%%%%%%%%%%%%%%%%
% Experimental Setup
%%%%%%%%%%%%%%%%%%%%%%%%%%%%%%
\subsection{Systematic Experiments Setup}

\begin{figure}[t]
\centering
\begin{minipage}{0.33\linewidth}
\centering
  \includegraphics[width=\linewidth]{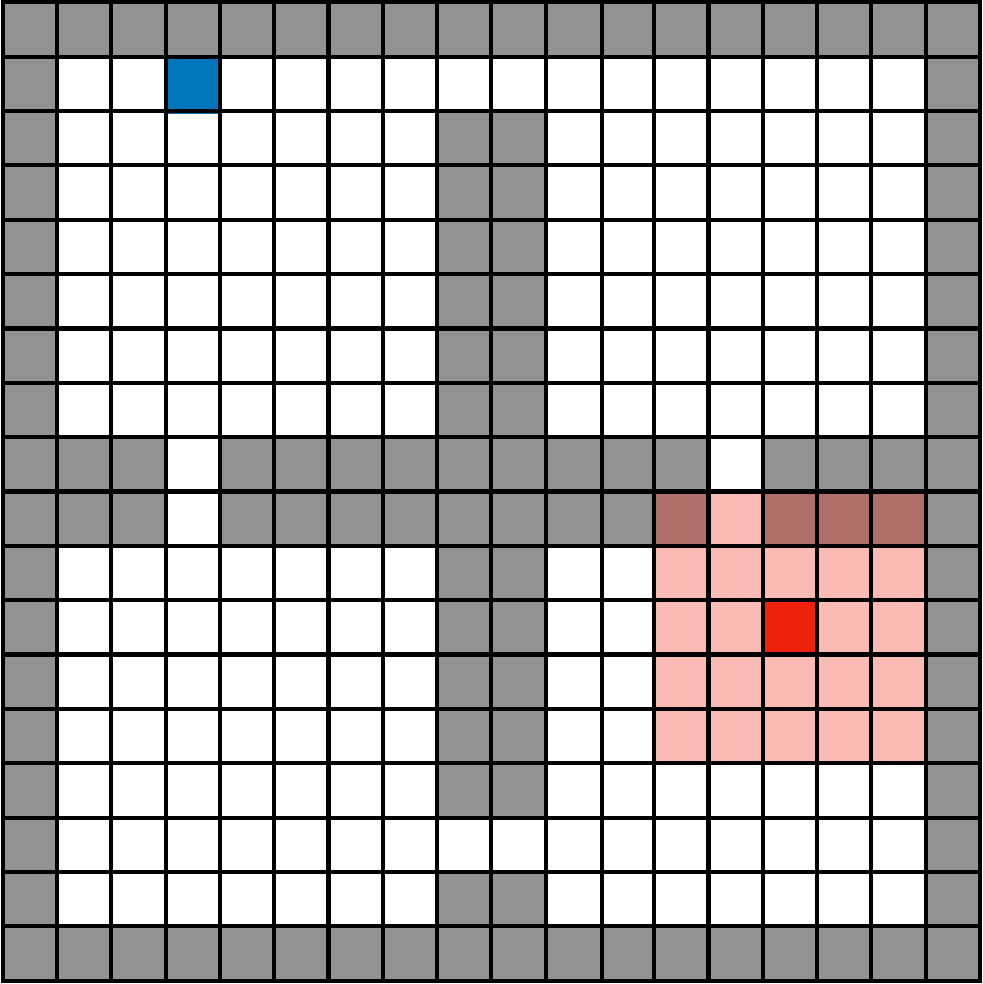}
  \small (a) P grid world
\end{minipage}
\hfill
\begin{minipage}{0.33\linewidth}
\centering
  \includegraphics[width=\linewidth]{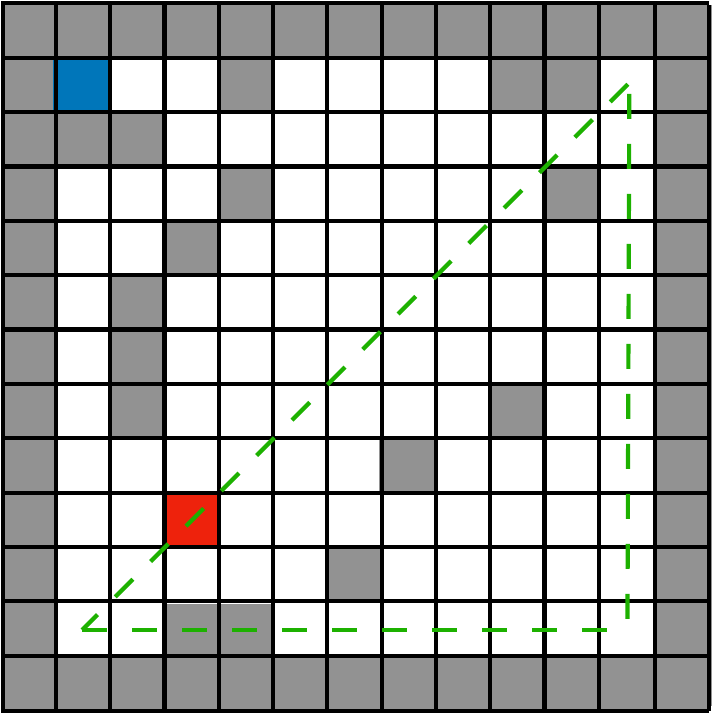}
  \small (b) F grid world
\end{minipage}
\hfill
\begin{minipage}{0.30\linewidth}
\centering
\begin{minipage}{0.49\linewidth}
\centering
  \includegraphics[width=\linewidth]{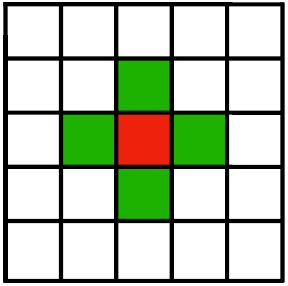}\\
  \small (c) 4-way
\end{minipage}
\hfill
\begin{minipage}{0.49\linewidth}
\centering
  \includegraphics[width=\linewidth]{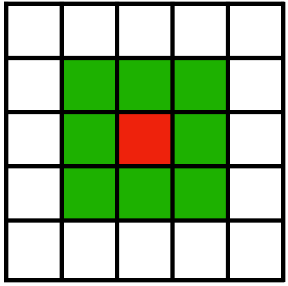}\\
  \small (d) king
\end{minipage}
\vskip 0.05in
\begin{minipage}{0.49\linewidth}
\centering
  \includegraphics[width=\linewidth]{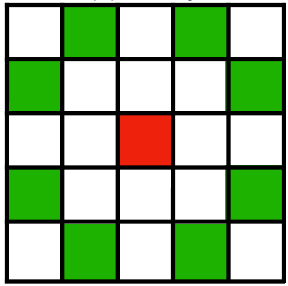}\\
  \small (e) knight
\end{minipage}
\hfill
\begin{minipage}{0.49\linewidth}
\centering
  \includegraphics[width=\linewidth]{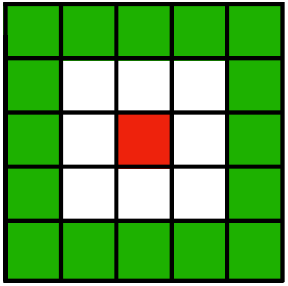}\\
  \small (f) 16-way
\end{minipage}
\vskip 0.1in
\end{minipage}
\caption{
(a-b) Example of an initial state of the environment with blue and red squares representing the goal and the agent, respectively. (a) In P grid world (a), with the $5\times5$ visible area around the agent highlighted. In F map (b), the agent is initially placed in one of the locations covered by green triangle. (c-f) Set of actions for different agents considered in our experiments. Green squares represent the possible locations after the move.
}
\label{fig:grid_and_actions}
\vskip -0.1in
\end{figure}

We consider two grid worlds with traps on the border (see Figure~\ref{fig:grid_and_actions}~(a-b)). These are navigation tasks -- the agent has to step on the goal to succeed. At the beginning of each episode, goal, agent and inner traps are randomly located. Hence, maps are different for each episode. More details on environments are presented in Appendix \ref{app:env_details}.

\paragraph{Action spaces} 
In our setup, the action space of the agent is isomorphic to the set of its moves. We consider four move styles (as illustrated in Figure~\ref{fig:grid_and_actions}~(c-f)): \emph{4-way} (up, down, left, right), \emph{king} (like king in the chess game), \emph{knight} (like knight in chess) and \emph{16-way} (has to move by 2 in one direction (horizontal or vertical), and 0, 1 or 2 in another one)\footnote{Knight and 16-way are able to ``jump'' over traps in F map.}. Note that the action space of 4-way and king agents (always move to adjacent locations) are disjoint from the other two agents (never move to adjacent squares). Also, the actions spaces of 4-way and knight are subsets of king and 16-way action spaces, respectively.

\paragraph{Experts} We consider two different types of rewards, that we call sparse and dense. The sparse version provides a signal at the termination of each episode only, giving either a reward $+1$ or $-1$ in case of success or failure of the task, respectively. As purposely designed, no agent is able to achieve the goal using the sparse reward function. We engineered the dense rewards such that all agents are able to succeed in this task and the four experts are trained in this way for each map.

%%%%%%%%%%%%%%%%%%%%%%%%%%%%%%
% Grid Worlds Results
%%%%%%%%%%%%%%%%%%%%%%%%%%%%%%
\subsection{Grid Worlds Results}
For each map we consider all possible expert-learner agent combinations, resulting in a total of 16 pairs. For each pair, we compare all methods proposed in Section~\ref{sec:method}.
Each experiment is executed five times (with different seeds) and results are shown with their mean and standard deviation over all trials. We consider a thousand observations (expert trajectories) for P map, and ten thousands of them for the F map. This number is about three order of magnitude lower than the number of trails required to train the expert agents with dense reward. We note that maps are randomly built for each episode -- therefore just a fraction of maps have the expert observation performed on them.

\newcommand{\showbar}[3]{
  \begin{minipage}{0.24\textwidth}
  \includegraphics[width=\textwidth]{figures/bar/#3/#1.png}
  \centering\small
  #2
  \end{minipage}}
  
\begin{figure*}[t]
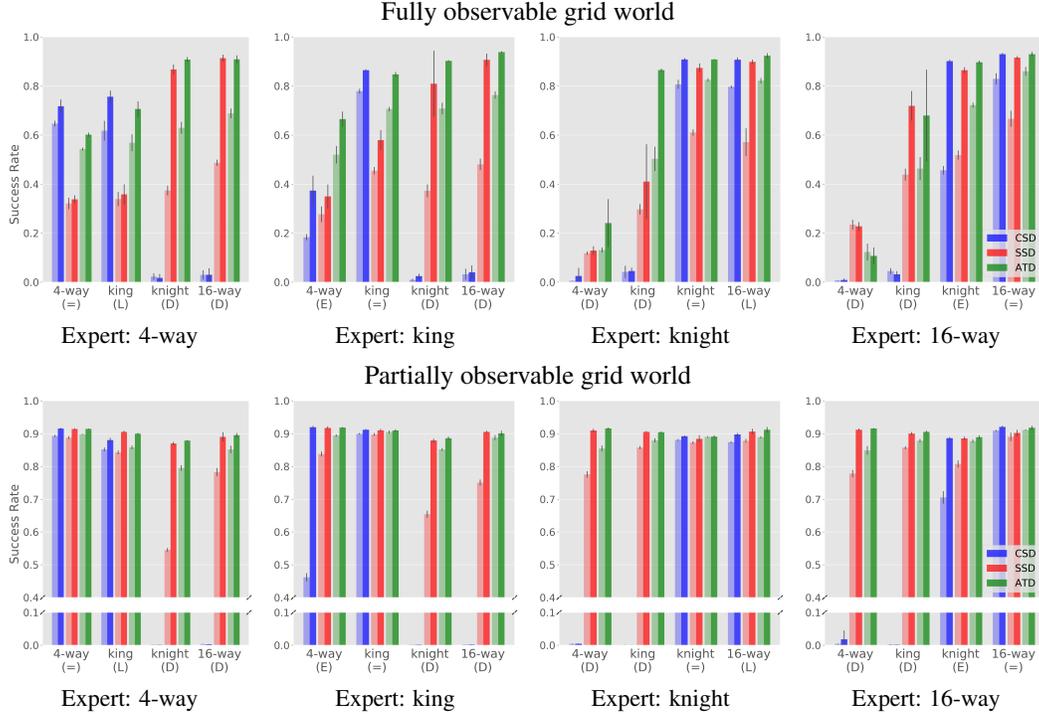
\centering
Fully observable grid world\\
\showbar{4-way}{Expert: 4-way}{mdp}
\showbar{king}{Expert: king}{mdp}
\showbar{knight}{Expert: knight}{mdp}
\showbar{16-way}{Expert: 16-way}{mdp}
\\
\vspace{0.2cm}
Partially observable grid world\\
\showbar{4-way}{Expert: 4-way}{pomdp}
\showbar{king}{Expert: king}{pomdp}
\showbar{knight}{Expert: knight}{pomdp}
\showbar{16-way}{Expert: 16-way}{pomdp}
\\
\caption{Comparison of RILO experiments with different imitation rewards and the effect of self-exploration on different expert-learner pairs. In the first row results for F map are presented (a-d) while results for P grid world are shown in the second row (e-h). Each chart represents how a given expert aids each learner, \emph{i.e.} learner move styles vary while expert is fixed. We show results in which agents use self-exploration (right, dark) or not (left, bright). Additionally, the relation between the learner and the expert is coded using one of four symbols: the same action spaces (=), disjoint action spaces (D), superior learner (L), or superior expert (E).}
\label{fig:ar-props}
\vskip -0.1in
\end{figure*}

We compare methods with different imitation reward strategies. Figure~\ref{fig:ar-props} shows results for all possible 16 expert-learner pairs considering three different imitation rewards: SSD, CSD and ATD. Experimental results on RTGD are very similar to (and implies in similar conclusion of) ATD. For this reason and for clarity, we remove them from the bar plots.

To disentangle the effect of self-exploration, we also show results where learners are trained with self-exploration (right, dark bar on the figure), dubbed SSD-SE, CSD-SE and ATD-SE, or without (left, bright bar on the figure). We consider models trained without self-exploration, especially CSD (a method similar to~\citet{torabi2018generative}), and SSD (similar to~\citet{merel2017learning}), as baseline methods. 

We notice that self-exploration significantly helps in the RILO setting. The learner policy is consistently better when given the opportunity to explore the environment without expert supervision. In very simple cases (for example the same action spaces for the expert and the learner), the effect of self-exploration usually reduces to a few percent points, although it still helps significantly.

\paragraph{Same action spaces} When the move styles are the same for the learner and the expert, $\mathcal{A}^e=\mathcal{A}^l$ (code (=) on Fig.~\ref{fig:ar-props}), the performance of CSD and ATD (and their counterparts with self-exploration) are similar. This means that using the consecutive states works well when the learner and the expert share the same action space. SSD, the second baseline, works well in same action spaces cases on P map, but it is significantly worse on F map. However, the difference in performance tends to diminish when the agent is trained in self-exploration mode (SSD-SE).

\paragraph{Disjoint action spaces} Consecutive states should not be used when learner and expert have disjoint actions spaces, $\mathcal{A}^e\cap\mathcal{A}^l=\emptyset$ (code (D) on Figure~\ref{fig:ar-props}). In all these cases (eight of them for each environment), CTD is not able to solve the task and the final success rate never exceeds 5\%. Even self-exploration is not able to give any improvement due to the poor success rate. On the other hand, with the use of self-exploration, SSD-SE and ATD-SE perform very well in disjoint cases, obtaining the best success rates. 

\paragraph{Superior learner} Next, we analyze the scenario where the learner has a superior set of actions, \emph{i.e.} $\mathcal{A}^e \subsetneq \mathcal{A}^l$ (code (L) on Figure~\ref{fig:ar-props}). It is, of course, not a disjoint case and all methods perform well. However, SSD and SSD-SE perform significantly worse on F map. Note that the superior learner can always imitate the expert (and limit itself to a smaller number of actions) so the discriminator is not able to observe that different actions may be performed. We noticed, however, that the self-exploration agents achieve the goal faster (in terms of number of steps) in this scenario. Hence, the learner is more likely to use actions that cannot be used by the expert but lead to better solution (note that due to the discount factor the solutions using less steps are preferable). In these four learner-expert scenarios (two per environment), the learner achieve the goal in about 15\% less steps when trained with self-exploration (when CSD-SE or ATD-SE strategies are used).

\begin{figure*}[t]
\centering
Fully observable grid world\\
\begin{minipage}{0.24\linewidth}
\centering
\includegraphics[width=1\linewidth]{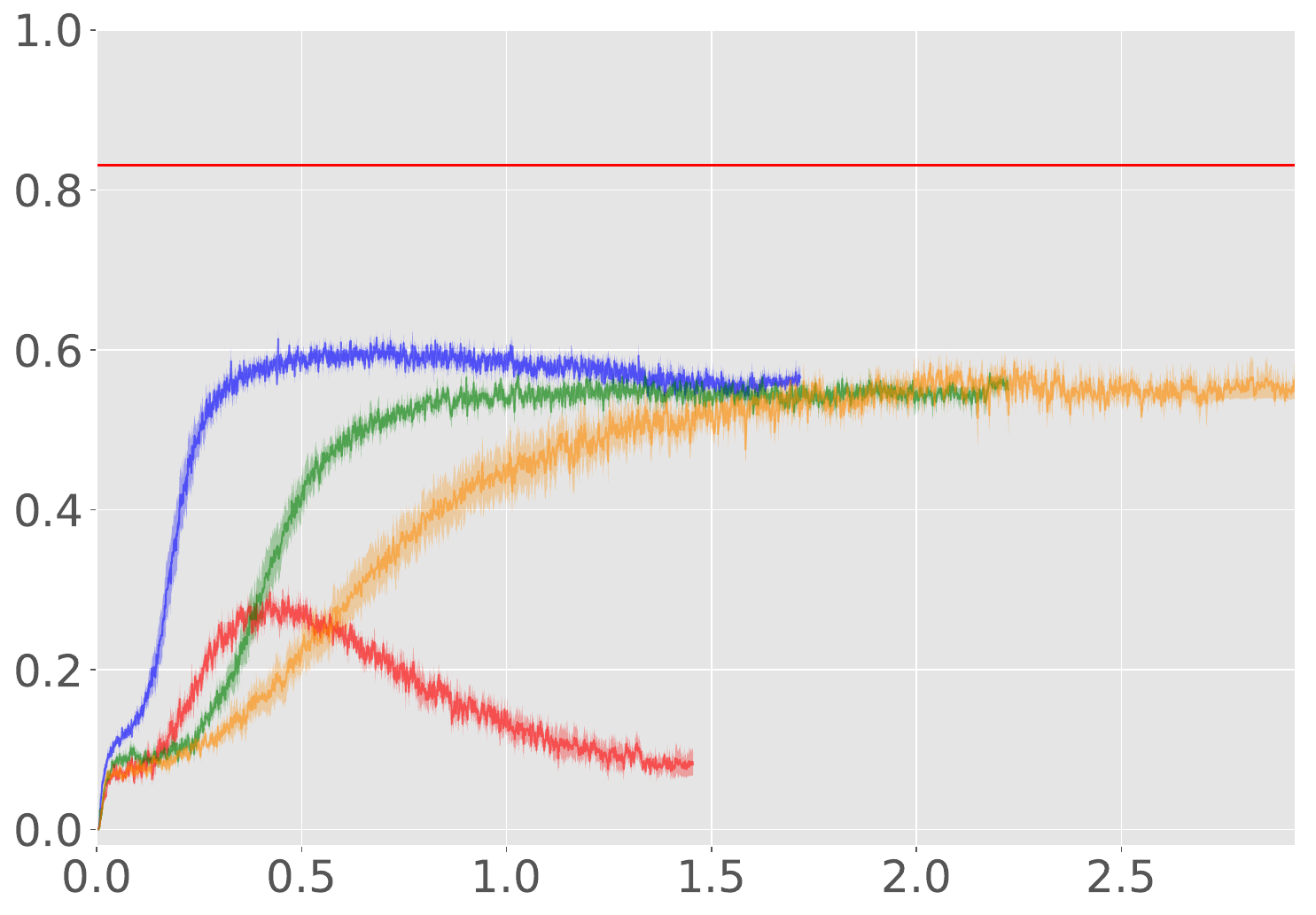}
\small (a) Learner: 4-way
\end{minipage}
%\hfill
\begin{minipage}{0.24\linewidth}
\centering
\includegraphics[width=1\linewidth]{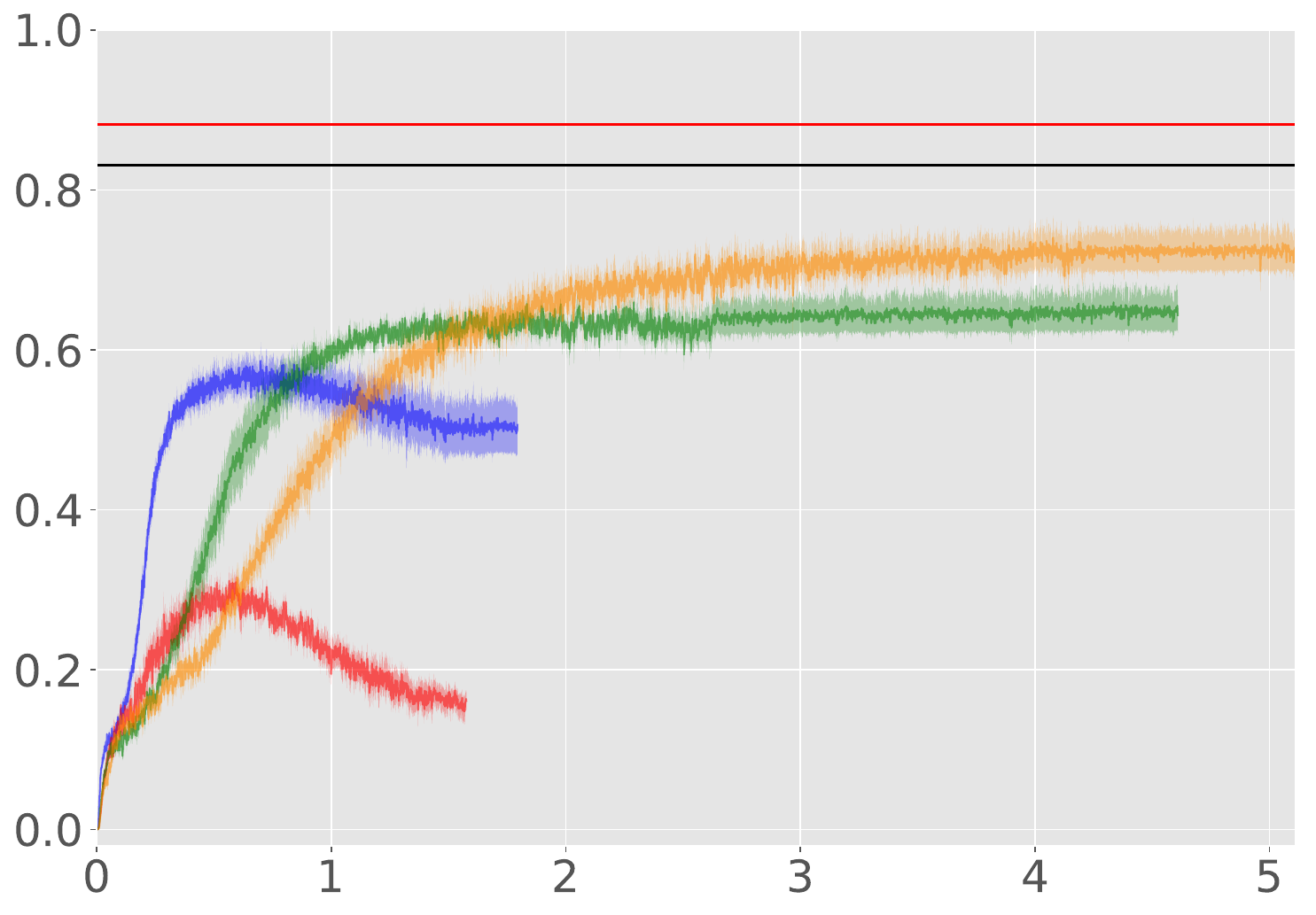}
\small (b) Learner: king
\end{minipage}
%\hfill
\begin{minipage}{0.24\linewidth}
\centering
\includegraphics[width=1\linewidth]{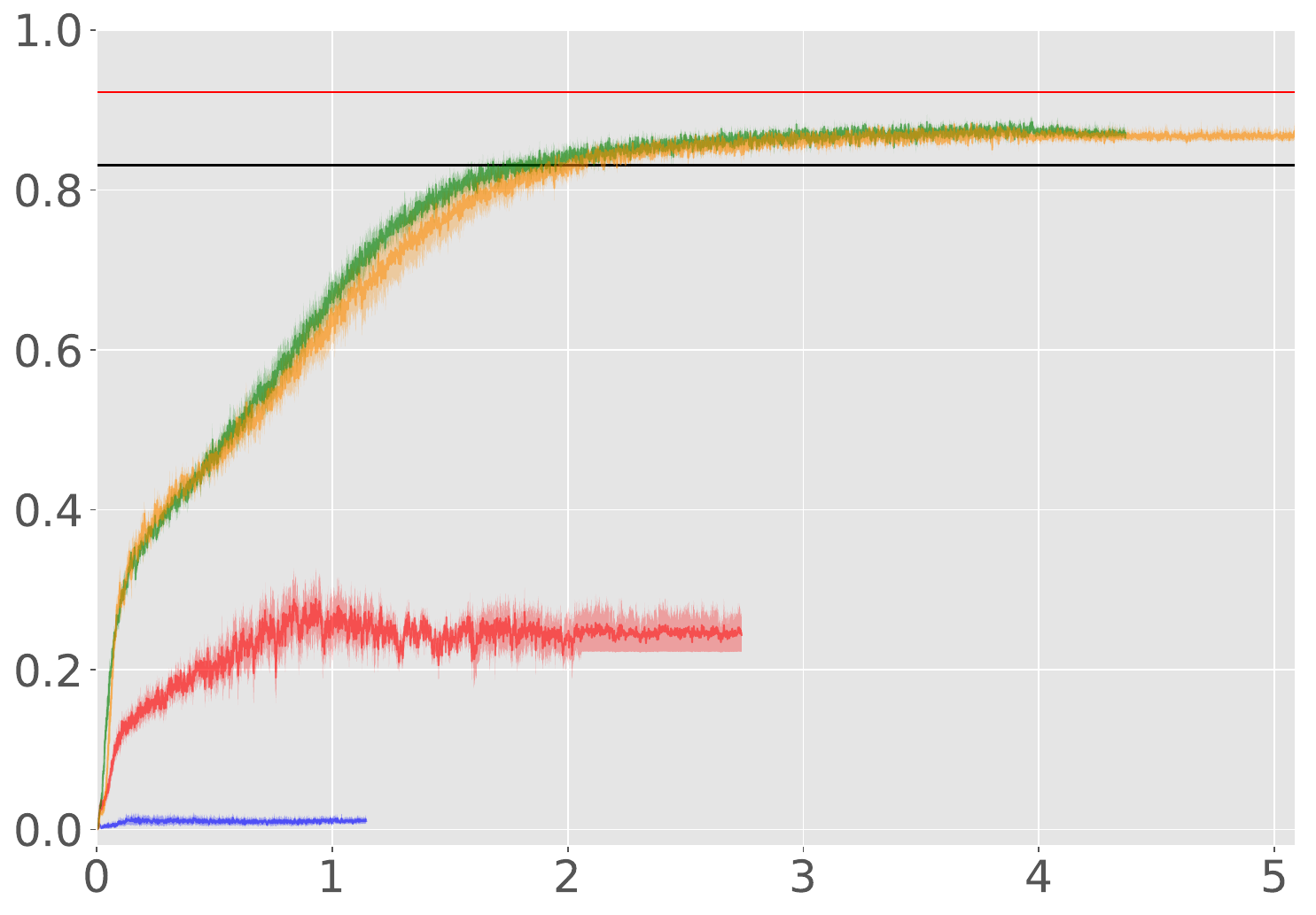}
\small (c) Learner: knight
\end{minipage}
%\hfill
\begin{minipage}{0.24\linewidth}
\centering
\includegraphics[width=1\linewidth]{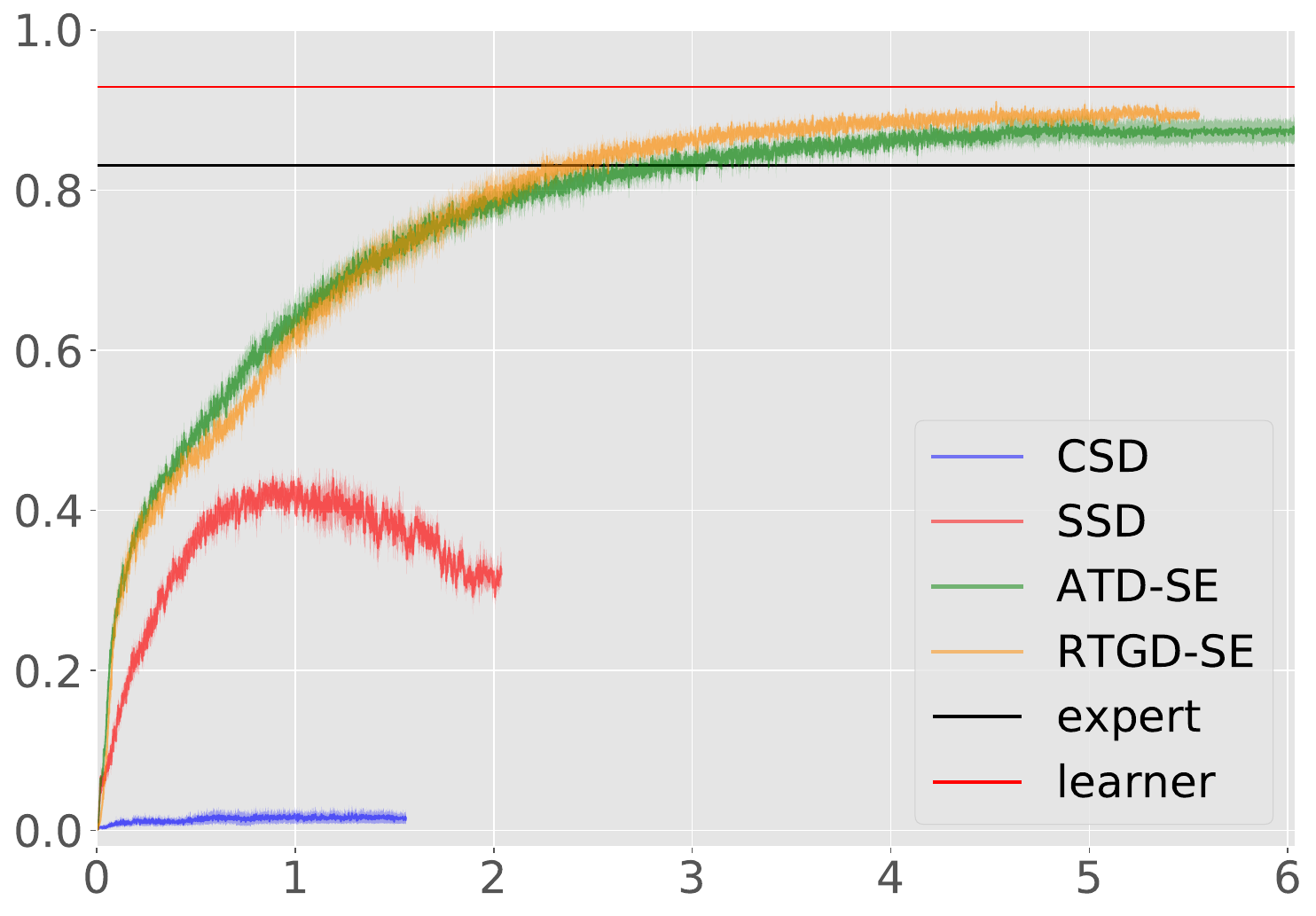}
\small (d) Learner: 16-way
\end{minipage}
\\
\vskip 0.2cm
Partially observable grid world\\
\begin{minipage}{0.24\linewidth}
\centering
\includegraphics[width=1\linewidth]{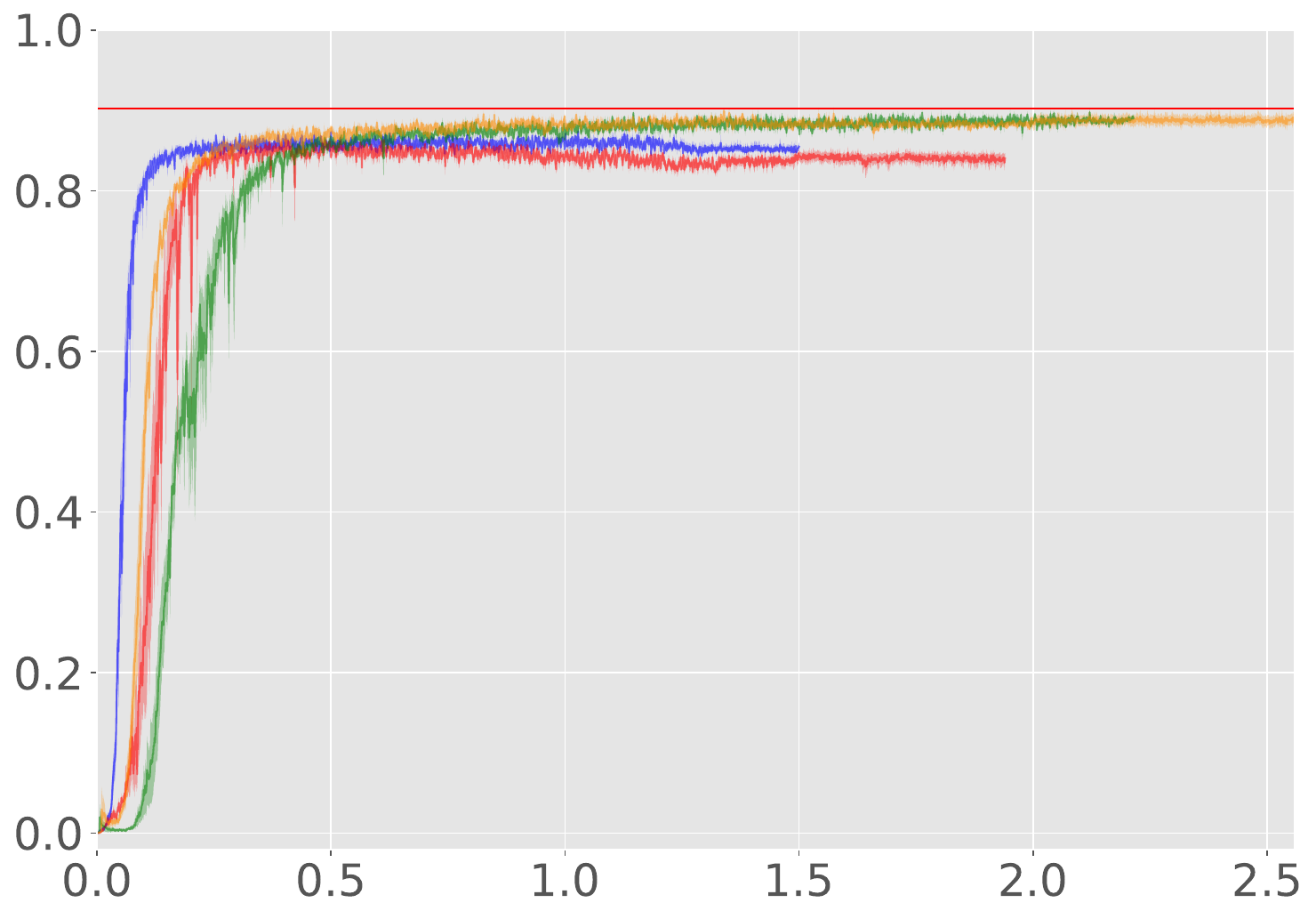}
\small (e) Learner: 4-way
\end{minipage}
%\hfill
\begin{minipage}{0.24\linewidth}
\centering
\includegraphics[width=1\linewidth]{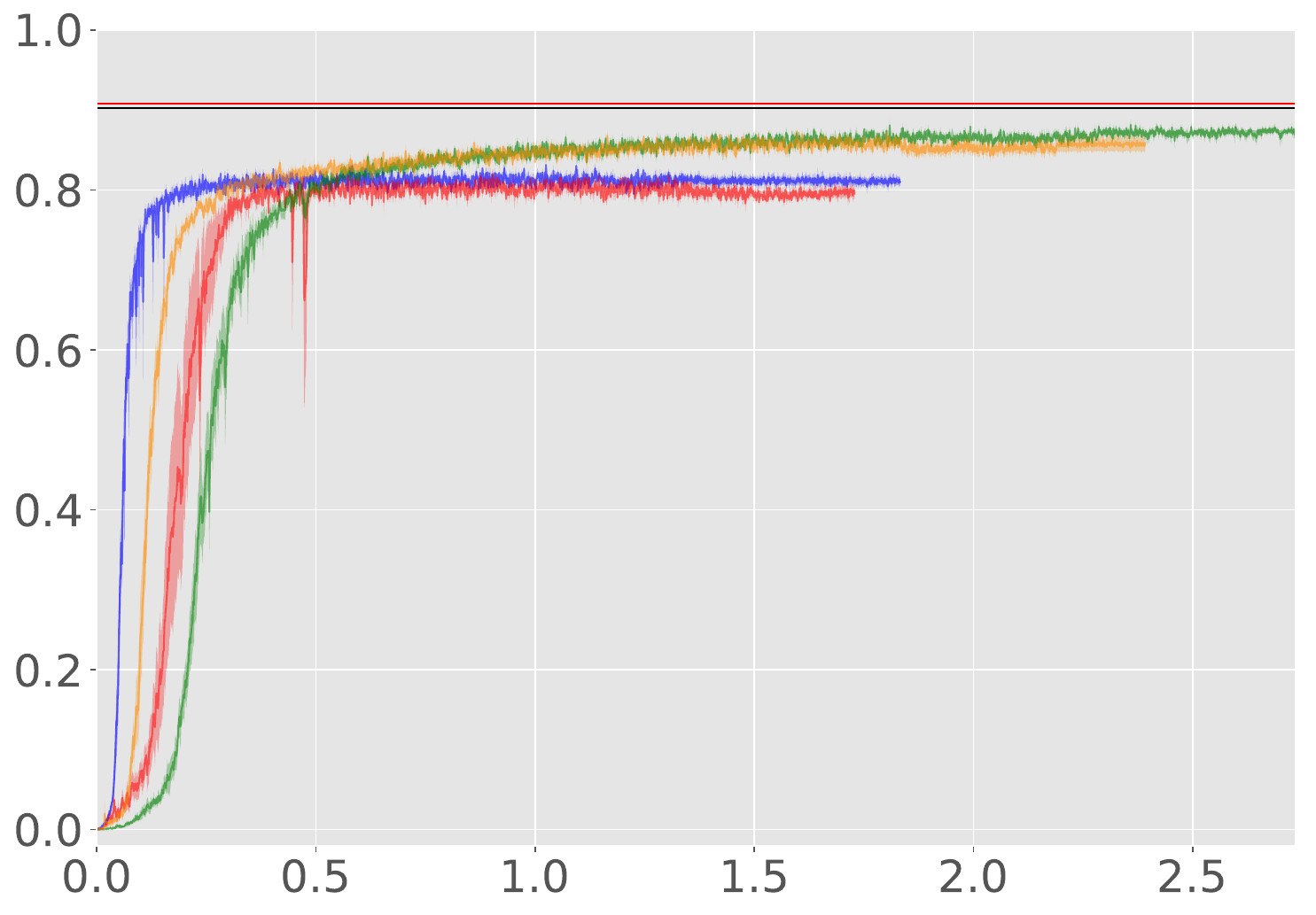}
\small (f) Learner: king
\end{minipage}
%\hfill
\begin{minipage}{0.24\linewidth}
\centering
\includegraphics[width=1\linewidth]{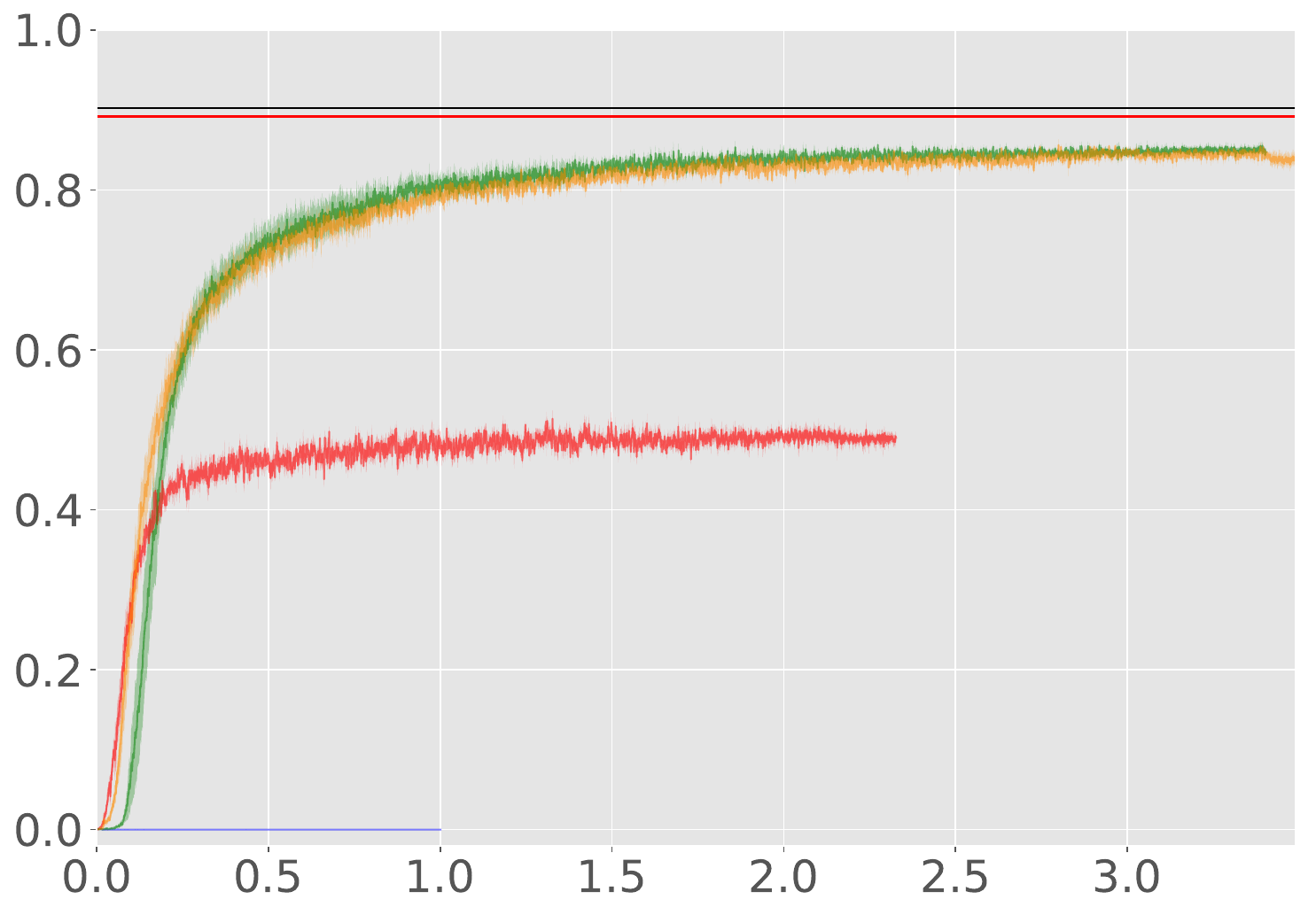}
\small (g) Learner: knight
\end{minipage}
%\hfill
\begin{minipage}{0.24\linewidth}
\centering
\includegraphics[width=1\linewidth]{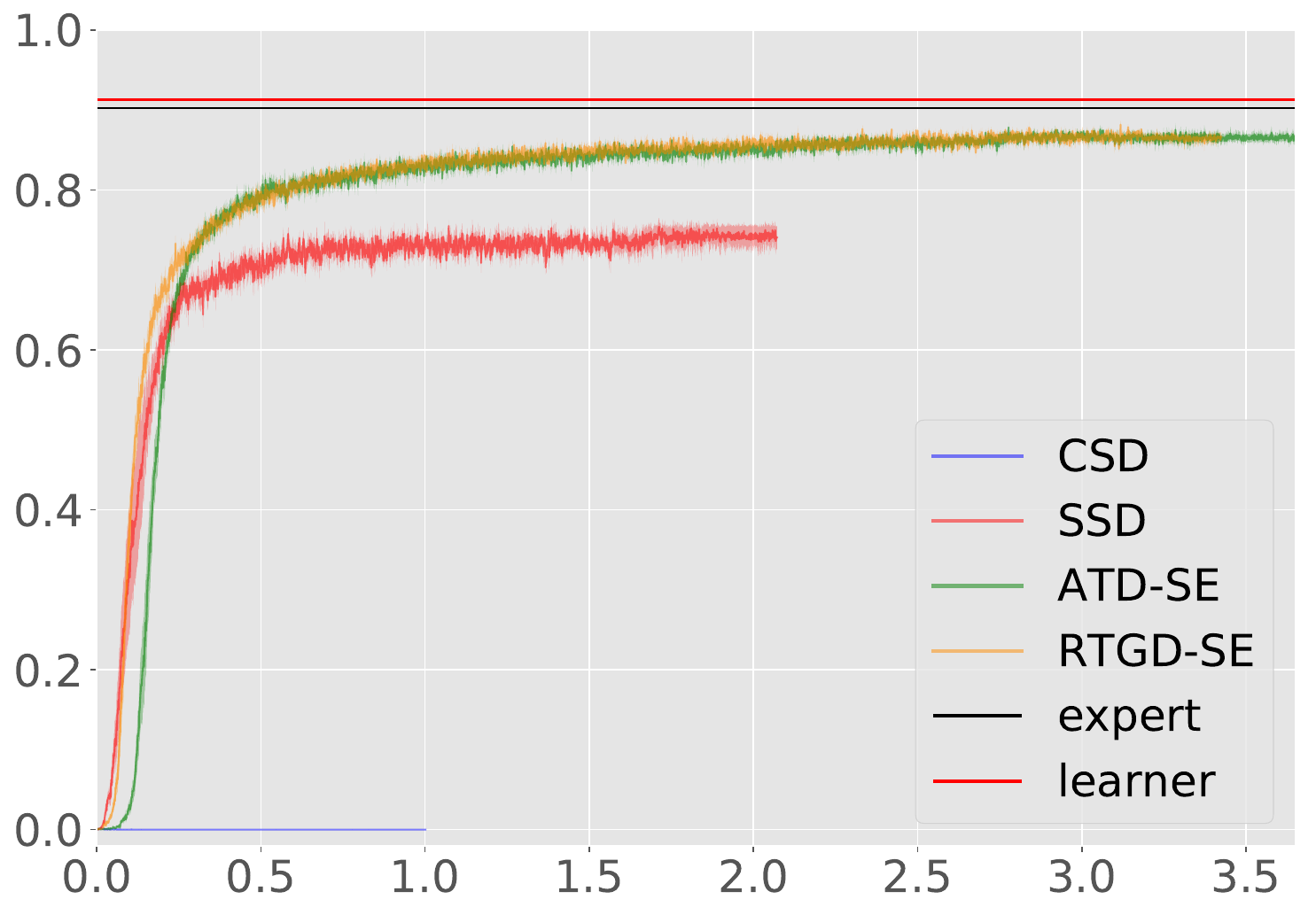}
\small (h) Learner: 16-way
\end{minipage}
\caption{Performance of algorithms (success rate) with respect to the number of iterations (in millions), assuming different learners. In all cases, the expert is the 4-way agent. See text for details and Appendix~\ref{app:plots} for higher resolution figures or other experts results.}
\label{fig:baseline_vs_ours}
\end{figure*}

% Comparison with baseline
\paragraph{Further comparison to baselines}
Figure~\ref{fig:baseline_vs_ours} compares the methods when 4-way expert is used along with all different learners (for both F grid world and P grid world). Results for other experts can be found in the Appendix~\ref{app:plots}. We present agents ATD-SE and RTGD-SE (our methods) and the baseline methods (CSD and SSD). We also included the performance of both agents trained with dense reward. In preliminary experiments, we observed that RTGD with minimal gap hyperparameter set to three works the best. This value is set fixed in all our experiments.

As shown in Figure~\ref{fig:baseline_vs_ours}, when expert's and learner's actions are disjoint (c-d, g-h) our methods usually reach significantly better performance and converge  faster. In these experiments on F map, our methods are able to outperform the expert, approaching the upper bound of the learner trained with dense reward. Again, the CSD performs well only when the learner's actions are the same or are a superset of the expert's actions (a-b, e-f). We can also see signs of overfitting on the SSD strategy trained on F map.

\paragraph{Observation limit}
The same set of 16 experiments were performed with smaller amounts of expert observation trajectories. When 500 trajectories are used for P environment (originally 1000 observations) the results just slightly deteriorate. When this number is further limited to 100 trajectories only, the positive effect of self-exploration is even more apparent. However, this reduction makes our problem much more challenging and only 12 out of 16 methods remain solved (by at least one approach). Using 50 trajectories is enough for only a few expert-learner pairs.

We also checked how different amount of observations limits influence results on F map (originally ten thousand of observations). We considered only one thousand expert observations. This reduction turns out to make the problem very challenging. Baseline methods (CSD and SSD) never bypass 50\% success rate, and usually score around 20\%. ATG and RTGD perform significantly better: a success rate above 50\% in 8 out of 16 expert-learner pairs (but achieves more than 90\% only in three cases). When self-exploration is used, both methods (ATG-SE and RTGD-SE) achieve success rate above 90\% in 8 out of the 16 setups.

At the first glance, it may be surprising that fully observable grid world requires more observations (and also the agents usually perform better in P map as compared to F map). However, the access to the full map and stochasticity of the maps make it much easier for the discriminator to remember maps and discriminate based on that. So,  while full observability makes the problem easier for the agent, the discriminator ``benefits'' more, making the training procedure loop more challenging. Additionally, LSTM agent is used to solve P map (some form of memory is needed to solve the task) as compared to non-recurrent architecture used in F map.

\paragraph{Coherence of results}
Results on both settings are coherent and justify the importance of self-exploration. The only significant difference in results is that the gap between SSD-SE and ATD-SE (or RTGD-SE) is smaller for P grid world. We hypothesize that this is due to the fact that the discriminator is more prone to overfitting (remember all expert trajectories) when given a full map (not just a small part as in P environment) and that problem is more severe when only a single state is given, not a random pair of states. To validate this hypothesis we trained models assuming unlimited observations from the expert for F map. In this case, the performance of SSD-SE approaches that of ATD-SE and RTGD-SE. This means that when given a massive amount of expert data, simply having the distribution of states is enough to infer the policy.

\paragraph{Comparison to pure IL}
We also checked how the presented strategies work in a pure IL setting, \emph{i.e.} when the sparse environment reward is not given. Not surprisingly, the performance of all method deteriorate. Our methods (RTGD and ATD) again perform better than the baseline methods. However, in pure IL setup the self-exploration cannot be applied and the differences in performance are smaller.

%%%%%%%%%%%%%%%%%%%%%%%%%%%%%%
% ViZDoom Results
%%%%%%%%%%%%%%%%%%%%%%%%%%%%%%

\subsection{ViZDoom Results}\label{subsec:vizdoom}

We also experiment on the ViZDoom engine to check how our results generalize to 3D POMDP setting. We chose \emph{my way home} map for our experiments (the version used in \cite{pathak2017curiosity}).  As in previous experiments, the expert is trained with dense rewards. All agents are trained from raw pixel frames (RGB). More details on the environment setup are presented in supplemental material.

\paragraph{Action spaces} The agent has access to three buttons: move forward, turning left or turning right.
The agent has three possible action spaces: \emph{single} (agent has access to one button per time unit), \emph{multi} (any number of buttons per time unit), \emph{right} (as multi but excluding button turning left). 
Due to computational limitations, we experiment with only one expert. We chose the \emph{single} action space for expert and consider all three possible options for learner.

\begin{figure*}[t]
\begin{center}
\centerline{\includegraphics[width=.8\linewidth]{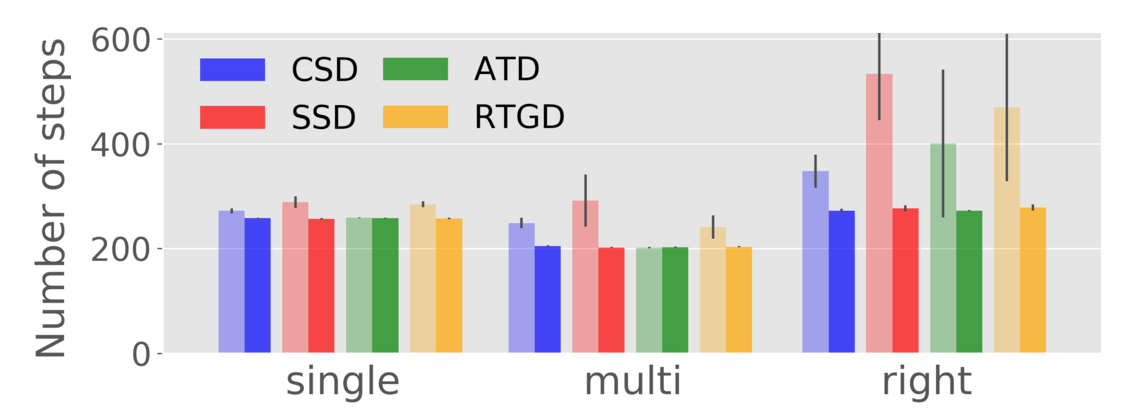}}
\caption{Comparison of RILO experiments with different imitation rewards and the effect of self-exploration for different learners. Expert is \emph{single} in all cases, hence the first set of experiments (\emph{single}) is with agent having the same action space, agent is superior for the middle one (\emph{multi}). The largest difference between action spaces is for the last case (\emph{right}).}
\label{fig:vizdoom}
\end{center}
\vskip -0.1in
\end{figure*}

We only consider 30 observation trajectories from the expert. Since all agents achieve nearly 100\% success rate, we found it to be more informative to compare the average number of steps needed to reach the goal (when the goal is not reached, the episode terminates after 2100 steps). Results are presented in Figure~\ref{fig:vizdoom}.

These results corroborate with findings of the previous subsection. Self-exploration always helps but the gap is larger when action spaces of learner and expert are different. 
For the same action spaces (\emph{single} on Figure~\ref{fig:vizdoom}) the self-exploration agent achieves the goal using 10\% less steps. The gap raises to around 20\% for the superior learner (\emph{multi}) and up to 50\% for \emph{right} agent, \emph{i.e.} the self-exploration makes agent solve the task twice as fast.
There does not seem to be any significant change between applying different imitation rewards considered (CSD, SSD, ATD or RTGD) when self-exploration is used which proves its generality.

%%%%%%%%%%%%%%%%%%%%%%%%%%%%%%%%%%%%%%%%%%%%%%%%%%%%%%%%%%%%%%%%%%%%%%%%%%%%%%%
% Conclusion
%%%%%%%%%%%%%%%%%%%%%%%%%%%%%%%%%%%%%%%%%%%%%%%%%%%%%%%%%%%%%%%%%%%%%%%%%%%%%%%
\section{Conclusion}\label{sec:conclusion}
In this paper, we show that by leveraging unshaped rewards from the environment, an agent is able to outperform standard state-only imitation learning. Our proposed method efficiently combines the sparse environment rewards with the standard imitation learning objective. Additionally, it adapts the agent's policy based on either mimicking expert's behavior or maximizing sparse rewards. It means that the agent can learn to behave differently when asked to imitate the expert or when it maximizes environment rewards only. We show experimentally that this approach achieves good performance over baselines in the RILO setting. Our method is especially well-suited when the actions of the trained agent differ from those of the expert. We also show that an agent trained with our approach can outperform the expert by using the sparse rewards in an optimized way.

\section*{Acknowledgments}

Konrad \.Zo\l{}na is supported by the National Science Center, Poland (2017/27/N/ST6/00828).

\small
\bibliographystyle{plainnat}
\bibliography{paper}

\begin{thebibliography}{32}
\providecommand{\natexlab}[1]{#1}
\providecommand{\url}[1]{\texttt{#1}}
\expandafter\ifx\csname urlstyle\endcsname\relax
  \providecommand{\doi}[1]{doi: #1}\else
  \providecommand{\doi}{doi: \begingroup \urlstyle{rm}\Url}\fi

\bibitem[Abbeel and Ng(2004)]{abbeel2004appren}
Pieter Abbeel and Andrew~Y. Ng.
\newblock Apprenticeship learning via inverse reinforcement learning.
\newblock In \emph{ICML}, 2004.

\bibitem[Aytar et~al.(2018)Aytar, Pfaff, Budden, Paine, Wang, and
  de~Freitas]{aytar2018playing}
Yusuf Aytar, Tobias Pfaff, David Budden, Tom~Le Paine, Ziyu Wang, and Nando
  de~Freitas.
\newblock Playing hard exploration games by watching youtube.
\newblock \emph{arXiv preprint arXiv:1805.11592}, 2018.

\bibitem[Bain and Sammut(1995)]{bainS95}
Michael Bain and Claude Sammut.
\newblock A framework for behavioural cloning.
\newblock In \emph{Machine Intelligence}, 1995.

\bibitem[Borsa et~al.(2017)Borsa, Piot, Munos, and
  Pietquin]{borsa2017observational}
Diana Borsa, Bilal Piot, R{\'e}mi Munos, and Olivier Pietquin.
\newblock Observational learning by reinforcement learning.
\newblock \emph{arXiv preprint arXiv:1706.06617}, 2017.

\bibitem[Duan et~al.(2017)Duan, Andrychowicz, Stadie, Ho, Schneider, Sutskever,
  Abbeel, and Zaremba]{duan2017one}
Yan Duan, Marcin Andrychowicz, Bradly Stadie, OpenAI~Jonathan Ho, Jonas
  Schneider, Ilya Sutskever, Pieter Abbeel, and Wojciech Zaremba.
\newblock One-shot imitation learning.
\newblock In \emph{NIPS}, 2017.

\bibitem[Gao et~al.(2018)Gao, Lin, Yu, Levine, Darrell,
  et~al.]{gao2018reinforcement}
Yang Gao, Ji~Lin, Fisher Yu, Sergey Levine, Trevor Darrell, et~al.
\newblock Reinforcement learning from imperfect demonstrations.
\newblock \emph{arXiv preprint arXiv:1802.05313}, 2018.

\bibitem[Goodfellow et~al.(2014)Goodfellow, Pouget-Abadie, Mirza, Xu,
  Warde-Farley, Ozair, Courville, and Bengio]{goodfellow2014gan}
Ian~J. Goodfellow, Jean Pouget-Abadie, Mehdi Mirza, Bing Xu, David
  Warde-Farley, Sherjil Ozair, Aaron Courville, and Yoshua Bengio.
\newblock Generative adversarial nets.
\newblock In \emph{NIPS}, 2014.

\bibitem[Gupta et~al.(2017)Gupta, Devin, Liu, Abbeel, and
  Levine]{gupta2017learning}
Abhishek Gupta, Coline Devin, YuXuan Liu, Pieter Abbeel, and Sergey Levine.
\newblock Learning invariant feature spaces to transfer skills with
  reinforcement learning.
\newblock \emph{arXiv preprint arXiv:1703.02949}, 2017.

\bibitem[Hester et~al.(2017)Hester, Vecerik, Pietquin, Lanctot, Schaul, Piot,
  Horgan, Quan, Sendonaris, Dulac-Arnold, et~al.]{hester2017deep}
Todd Hester, Matej Vecerik, Olivier Pietquin, Marc Lanctot, Tom Schaul, Bilal
  Piot, Dan Horgan, John Quan, Andrew Sendonaris, Gabriel Dulac-Arnold, et~al.
\newblock Deep q-learning from demonstrations.
\newblock \emph{arXiv preprint arXiv:1704.03732}, 2017.

\bibitem[Heyes(2010)]{heyes2010mirror}
Cecilia Heyes.
\newblock Where do mirror neurons come from?
\newblock \emph{Neuroscience \& Biobehavioral Reviews}, 2010.

\bibitem[Ho and Ermon(2016)]{ho2016generative}
Jonathan Ho and Stefano Ermon.
\newblock Generative adversarial imitation learning.
\newblock In \emph{NIPS}, 2016.

\bibitem[Jones(2009)]{jones2009development}
Susan~S Jones.
\newblock The development of imitation in infancy.
\newblock \emph{Philosophical Transactions of the Royal Society of London B:
  Biological Sciences}, 2009.

\bibitem[Kang et~al.(2018)Kang, Jie, and Feng]{kang2018policy}
Bingyi Kang, Zequn Jie, and Jiashi Feng.
\newblock Policy optimization with demonstrations.
\newblock In \emph{ICML}, 2018.

\bibitem[Kempka et~al.(2016)Kempka, Wydmuch, Runc, Toczek, and
  Jaskowski]{KempkaWRTJ16}
Michal Kempka, Marek Wydmuch, Grzegorz Runc, Jakub Toczek, and Wojciech
  Jaskowski.
\newblock Vizdoom: {A} doom-based {AI} research platform for visual
  reinforcement learning.
\newblock \emph{arXiv preprint arXiv:1605.02097}, 2016.

\bibitem[Kimura et~al.(2018)Kimura, Chaudhury, Tachibana, and
  Dasgupta]{kimura2018internal}
Daiki Kimura, Subhajit Chaudhury, Ryuki Tachibana, and Sakyasingha Dasgupta.
\newblock Internal model from observations for reward shaping.
\newblock \emph{arXiv preprint arXiv:1806.01267}, 2018.

\bibitem[Kingma and Ba(2014)]{kingmaB14adam}
Diederik~P. Kingma and Jimmy Ba.
\newblock Adam: A method for stochastic optimization.
\newblock \emph{ICLR}, 2014.

\bibitem[Li et~al.(2017)Li, Song, and Ermon]{li2017infogail}
Yunzhu Li, Jiaming Song, and Stefano Ermon.
\newblock Infogail: Interpretable imitation learning from visual
  demonstrations.
\newblock In \emph{NIPS}, 2017.

\bibitem[Liu et~al.(2017)Liu, Gupta, Abbeel, and Levine]{liu2017imitation}
YuXuan Liu, Abhishek Gupta, Pieter Abbeel, and Sergey Levine.
\newblock Imitation from observation: Learning to imitate behaviors from raw
  video via context translation.
\newblock \emph{arXiv preprint arXiv:1707.03374}, 2017.

\bibitem[McFarland(1999)]{david1999animal}
David McFarland.
\newblock \emph{Animal Behaviour: Psychobiology, Ethology and Evolution}.
\newblock 1999.

\bibitem[Merel et~al.(2017)Merel, Tassa, Srinivasan, Lemmon, Wang, Wayne, and
  Heess]{merel2017learning}
Josh Merel, Yuval Tassa, Sriram Srinivasan, Jay Lemmon, Ziyu Wang, Greg Wayne,
  and Nicolas Heess.
\newblock Learning human behaviors from motion capture by adversarial
  imitation.
\newblock \emph{arXiv preprint arXiv:1707.02201}, 2017.

\bibitem[Mnih et~al.(2016)Mnih, Badia, Mirza, Graves, Harley, Lillicrap,
  Silver, and Kavukcuoglu]{mnih16a3c}
Volodymyr Mnih, Adri\`{a}~Puigdom\`{e}nech Badia, Mehdi Mirza, Alex Graves, Tim
  Harley, Timothy~P. Lillicrap, David Silver, and Koray Kavukcuoglu.
\newblock Asynchronous methods for deep reinforcement learning.
\newblock In \emph{ICML}, 2016.

\bibitem[Nair et~al.(2017)Nair, McGrew, Andrychowicz, Zaremba, and
  Abbeel]{nair2017overcoming}
Ashvin Nair, Bob McGrew, Marcin Andrychowicz, Wojciech Zaremba, and Pieter
  Abbeel.
\newblock Overcoming exploration in reinforcement learning with demonstrations.
\newblock \emph{arXiv preprint arXiv:1709.10089}, 2017.

\bibitem[Ng and Russell(2000)]{ng2000ilr}
Andrew~Y. Ng and Stuart~J. Russell.
\newblock Algorithms for inverse reinforcement learning.
\newblock In \emph{ICML}, 2000.

\bibitem[Pathak et~al.(2017)Pathak, Agrawal, Efros, and
  Darrell]{pathak2017curiosity}
Deepak Pathak, Pulkit Agrawal, Alexei~A Efros, and Trevor Darrell.
\newblock Curiosity-driven exploration by self-supervised prediction.
\newblock In \emph{ICML}, 2017.

\bibitem[Pathak et~al.(2018)Pathak, Mahmoudieh, Luo, Agrawal, Chen, Shentu,
  Shelhamer, Malik, Efros, and Darrell]{pathak2018zero}
Deepak Pathak, Parsa Mahmoudieh, Guanghao Luo, Pulkit Agrawal, Dian Chen, Yide
  Shentu, Evan Shelhamer, Jitendra Malik, Alexei~A Efros, and Trevor Darrell.
\newblock Zero-shot visual imitation.
\newblock In \emph{ICLR}, 2018.

\bibitem[Pomerleau(1989)]{pomerleau89alvinn}
Dean~A. Pomerleau.
\newblock Alvinn: An autonomous land vehicle in a neural network.
\newblock In \emph{NIPS}, 1989.

\bibitem[Ratliff et~al.(2007)Ratliff, Bagnell, and Srinivasa]{ratliffBS07}
Nathan~D. Ratliff, James~A. Bagnell, and Siddhartha~S. Srinivasa.
\newblock Imitation learning for locomotion and manipulation.
\newblock In \emph{Humanoids}, 2007.

\bibitem[Stadie et~al.(2017)Stadie, Abbeel, and Sutskever]{stadie2017third}
Bradly~C Stadie, Pieter Abbeel, and Ilya Sutskever.
\newblock Third-person imitation learning.
\newblock \emph{arXiv preprint arXiv:1703.01703}, 2017.

\bibitem[Torabi et~al.(2018{\natexlab{a}})Torabi, Warnell, and
  Stone]{torabi2018behavioral}
Faraz Torabi, Garrett Warnell, and Peter Stone.
\newblock Behavioral cloning from observation.
\newblock \emph{arXiv preprint arXiv:1805.01954}, 2018{\natexlab{a}}.

\bibitem[Torabi et~al.(2018{\natexlab{b}})Torabi, Warnell, and
  Stone]{torabi2018generative}
Faraz Torabi, Garrett Warnell, and Peter Stone.
\newblock Generative adversarial imitation from observation.
\newblock \emph{arXiv preprint arXiv:1807.06158}, 2018{\natexlab{b}}.

\bibitem[Vecer{\'\i}k et~al.(2017)Vecer{\'\i}k, Hester, Scholz, Wang, Pietquin,
  Piot, Heess, Roth{\"o}rl, Lampe, and Riedmiller]{vecerik2017leveraging}
Matej Vecer{\'\i}k, Todd Hester, Jonathan Scholz, Fumin Wang, Olivier Pietquin,
  Bilal Piot, Nicolas Heess, Thomas Roth{\"o}rl, Thomas Lampe, and Martin~A
  Riedmiller.
\newblock Leveraging demonstrations for deep reinforcement learning on robotics
  problems with sparse rewards.
\newblock \emph{CoRR, abs/1707.08817}, 2017.

\bibitem[Zhu et~al.(2018)Zhu, Wang, Merel, Rusu, Erez, Cabi, Tunyasuvunakool,
  Kram{\'a}r, Hadsell, de~Freitas, et~al.]{zhu2018reinforcement}
Yuke Zhu, Ziyu Wang, Josh Merel, Andrei Rusu, Tom Erez, Serkan Cabi, Saran
  Tunyasuvunakool, J{\'a}nos Kram{\'a}r, Raia Hadsell, Nando de~Freitas, et~al.
\newblock Reinforcement and imitation learning for diverse visuomotor skills.
\newblock \emph{arXiv preprint arXiv:1802.09564}, 2018.

\end{thebibliography}


%\newpage
% { } $ $ { }
\newpage
\appendix

\section{Environments Details}\label{app:env_details}
\subsection{Grid Worlds}

We consider two grid worlds with traps on the border (see Figure~\ref{fig:grid_and_actions}~(a-b)). At the beginning of each episode the goal is randomly (uniformly) located (for P map we restrict the target location to be one of the 4 corners). The agent's initial location depends on both the goal and the map. For the P environment, the agent is always placed in different room, and for F map the agent is placed in one of the locations covered by a triangle made out of the three left corners (a total of $30$ possible initial locations per goal position -- dashed lines on Figure~\ref{fig:grid_and_actions}~(b)). On F map each of the squares (not taken by the goal or the agent) has $15\%$ chances of being a trap (gray squares on figure) while for P grid world traps are fixed to create 4 rooms (although the passages are randomly placed for each episode). If any agent steps on a trap, the episode is terminated with a final reward of $-1$. The episode is also terminated, if the agent performs its $51^{th}$ (or $26^{th}$ for F map) action. All other rewards are zero unless the agent steps on the goal which gives reward of $+1$ and also terminates the game. In all experiments, we use discount factor $\gamma=0.9$ and exploration rate $\epsilon=0.05$. Note that the maps are different for each episode.

\subsection{ViZDoom}

We picked \emph{my way home} map where the objective is to find and collect the vest. Hence, it is a navigation task. We used versions of the map presented in \citep{pathak2017curiosity}. Expert was trained using ``dense'' version of the map while the ``sparse'' version of the map was used for all imitation experiments. In that version the starting location is far from the goal which makes vanilla A3C algorithms not able to solve the task \citep{pathak2017curiosity}. We obtained the same results and confirm that with our implementation.

We used all default values for environment variables (proided in the original repository \citep{KempkaWRTJ16}) such as timeout and discount factor.

\section{Implementation Details}\label{app:implementation}

As mentioned in Subsection~\ref{overview}, our method is composed of two trainable components, a policy network $\pi_{\theta}:\mathcal{S}\to\mathcal{A}^l$ and a discriminator $D_{\phi}:\mathcal{S}\times\mathcal{S}\to[0,1]$. 
Both functions are parameterized by neural networks. In this section we present architectures for these networks and describe training hyper-parameters.

Source code that provide will be released upon acceptance and we believe that it is the best way to check all additional implementation details that can not be shared here.

\subsection{Grid Worlds}

Policy network and a discriminator have a state encoder that has identical architectures but with different set of weights. All weights are optimized using Adam~\citep{kingmaB14adam} with a learning rate of $10^{-4}$.
The training procedure is terminated after one million of episodes without significant performance improvement (defined to be $1\%$ average success rate increase).

\paragraph{State encoder} The encoder receives as input the grid world encoded as a matrix (for P map only a $5\times5$ subgrid around the agent is taken while the full $13\times13$ grid is visible in F environment) with $4$ possible values: $3$ for a goal, $2$ for an agent, $-1$ for all traps and walls and $0$ otherwise) and a few additional features\footnote{These features are two floats $(x, y) \in (0,1)^2$ encoding agent position. Additionally, as described before, the learner trained with self-exploration is given a binary variable indicating the nature of the reward that will be used.}. The map is processed by 5-layer CNN with kernel size $3$ and residual connections. Then, it is flattened and concatenated with additional features which constitutes the final state encoding.

\paragraph{Policy network} We use (synchronous) advantage actor-critic (A2C) algorithm~\citep{mnih16a3c} to optmize the policy.
The policy network encodes the state and then the final transformation is applied to obtain $(k+1)$ dimensional vector ($k$ possible actions, modeled as a probability with softmax, and one dimension to represent the value-function). The type of the final transformation depends on the environment. It is fully connected (with ReLU) for fully observable grid world and LSTM for partially observable grid world (since the memory is needed to perform the task well).

\paragraph{Discriminator} The discriminator encodes both input states separately (using the same state encoder) that are next separately inputted to $2$-layer MLP with $256$ hidden units (both layers) and ReLU activation function. The difference of the two encoded states are then fed to two fully-connected layers, with outputs sized 256 and 1, respectively. The final output is transformed into a probability with a sigmoid function.

Note that the computation time for the final state tuple is negligible compared to the computation for state encoding. As a result, ATD does not carries a heavy computation burden in case of relatively small trajectories. In case of very long trajectories, however, a fixed number of random pairs should be considered.

\paragraph{Imitation rewards normalization}
We used fixed value of $- \log 0.5$ that was subtracted to normalize imitation rewards, as described in Subsection \ref{subsec:normalization}.

\subsection{ViZDoom}

We used policy network architecture and training hyper-parameters for asynchronous advantage actor critic (A3C) as in \citep{pathak2017curiosity}.
The training procedure is terminated after 3 days of training using 20 CPU cores.

\paragraph{Discriminator} The discriminator encodes both input states separately (using simple 3-layer CNN network) that are next separately inputted to $2$-layer MLP with $64$ hidden units (both layers) and ReLU activation function. The difference of the two encoded states are then fed to two fully-connected layers, with outputs sized 64 and 1, respectively. The final output is transformed into a probability with a sigmoid function. Adam~\citep{kingmaB14adam} with a learning rate of $10^{-4}$ was used to train discriminator.

\paragraph{Imitation rewards normalization}
We used dynamic value computed as average of unnormalized imitation rewards for each batch to normalize them, as described in Subsection \ref{subsec:normalization}.

\section{Additional Plots}\label{app:plots}
Similar to Figure~\ref{fig:baseline_vs_ours}, we show performance of all algorithms with respect to the number of iterations  on Figures~\ref{fig:appendix1}-\ref{fig:appendix4}. Each plot assumes different expert (see caption) and presents result for all four learners.

\newcommand{\showplott}[2]{
  \begin{minipage}{0.4\linewidth}
  \centering
  \includegraphics[width=1\linewidth]{figures/evolution/#2/e_#1_l_4-way-plot.pdf}
  \small Learner: 4-way
  \end{minipage}
  \begin{minipage}{0.4\linewidth}
  \centering
  \includegraphics[width=1\linewidth]{figures/evolution/#2/e_#1_l_king-plot.pdf}
  \small Learner: king
  \end{minipage}

  \begin{minipage}{0.4\linewidth}
  \centering
  \includegraphics[width=1\linewidth]{figures/evolution/#2/e_#1_l_knight-plot.pdf}
  \small Learner: knight
  \end{minipage}
  \begin{minipage}{0.4\linewidth}
  \centering
  \includegraphics[width=1\linewidth]{figures/evolution/#2/e_#1_l_16-way-plot.pdf}
  \small Learner: 16-way
  \end{minipage}
  }

\begin{figure*}[t]
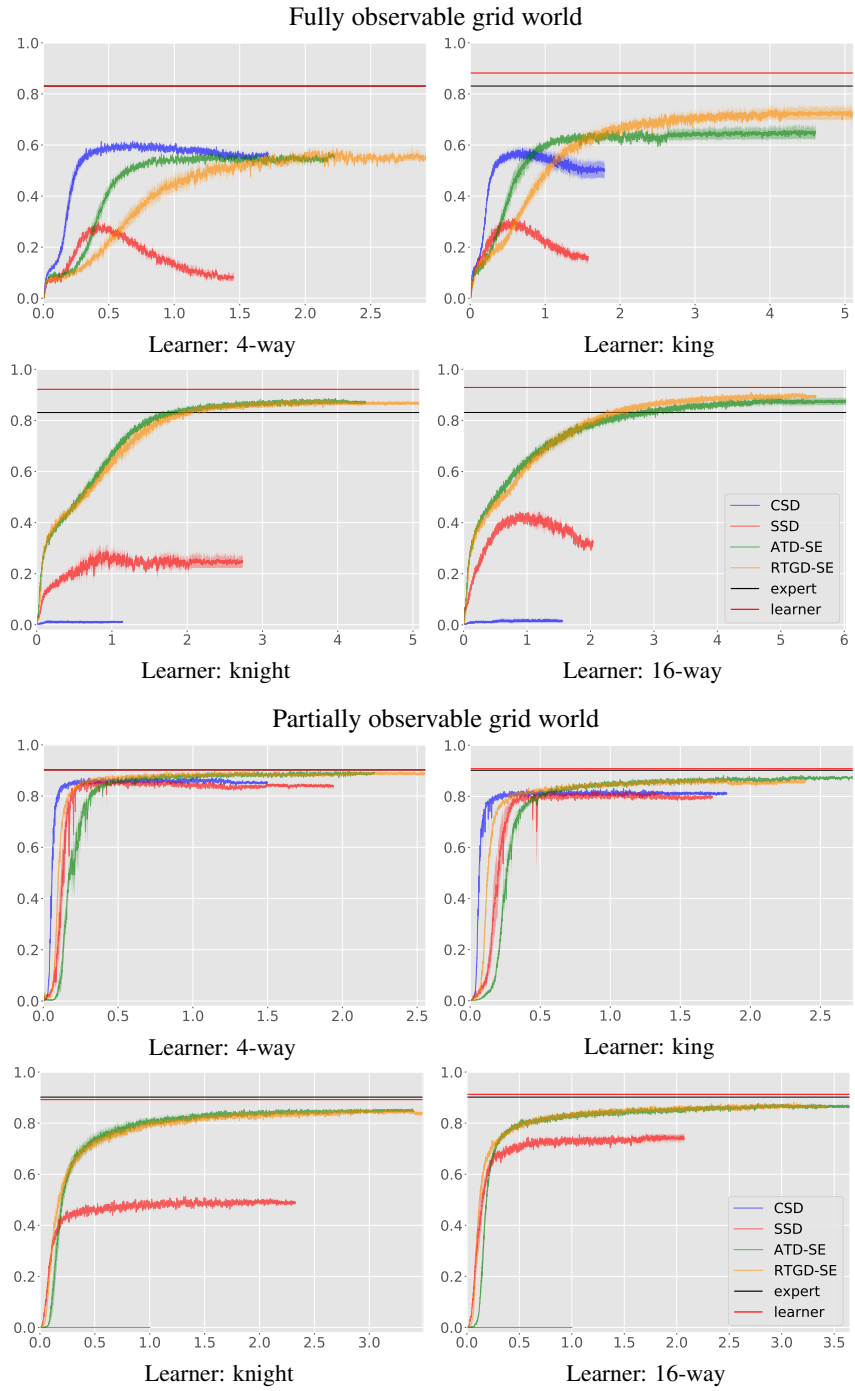
\centering
Fully observable grid world\\
\showplott{4-way}{mdp}
\vspace{0.3cm}\\
Partially observable grid world\\
\showplott{4-way}{pomdp}
\caption{Expert: 4-way}
\label{fig:appendix1}
\end{figure*}

\begin{figure*}[t]
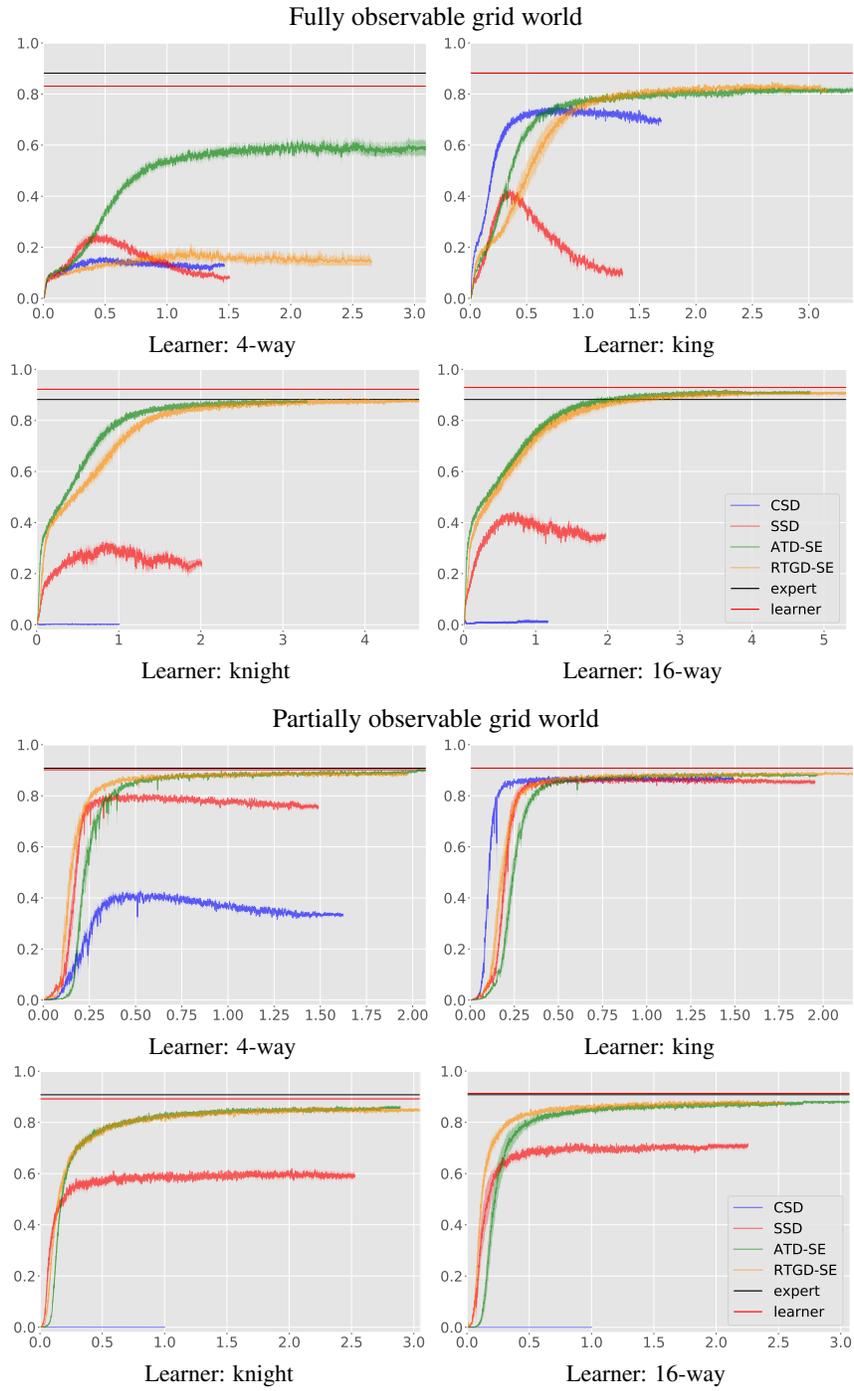
\centering
Fully observable grid world\\
\showplott{king}{mdp}
\vspace{0.3cm}\\
Partially observable grid world\\
\showplott{king}{pomdp}
\caption{Expert: king}
\label{fig:appendix2}
\end{figure*}

\begin{figure*}[t]
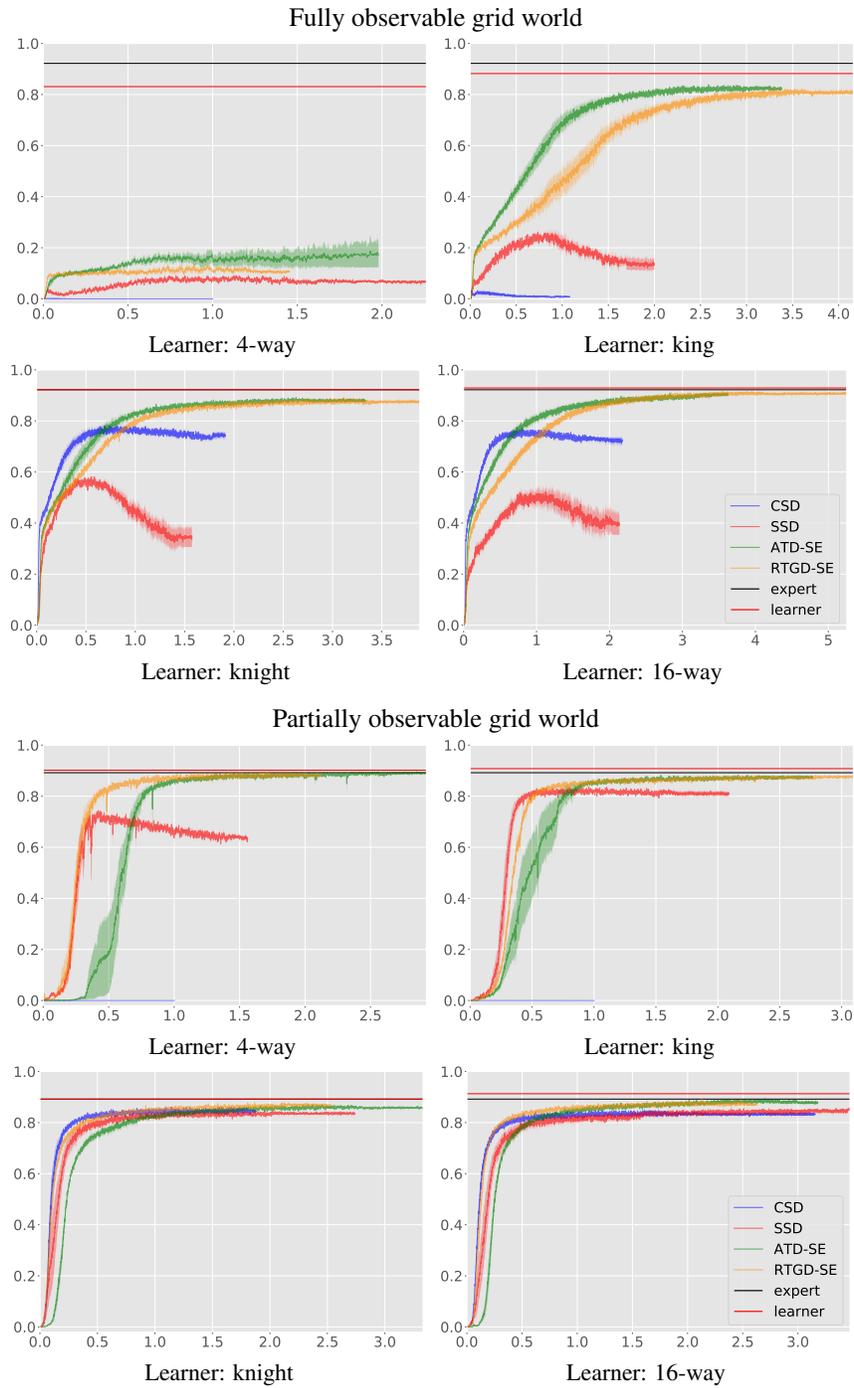
\centering
Fully observable grid world\\
\showplott{knight}{mdp}
\vspace{0.3cm}\\
Partially observable grid world\\
\showplott{knight}{pomdp}
\caption{Expert: knight}
\label{fig:appendix3}
\end{figure*}

\begin{figure*}[t]
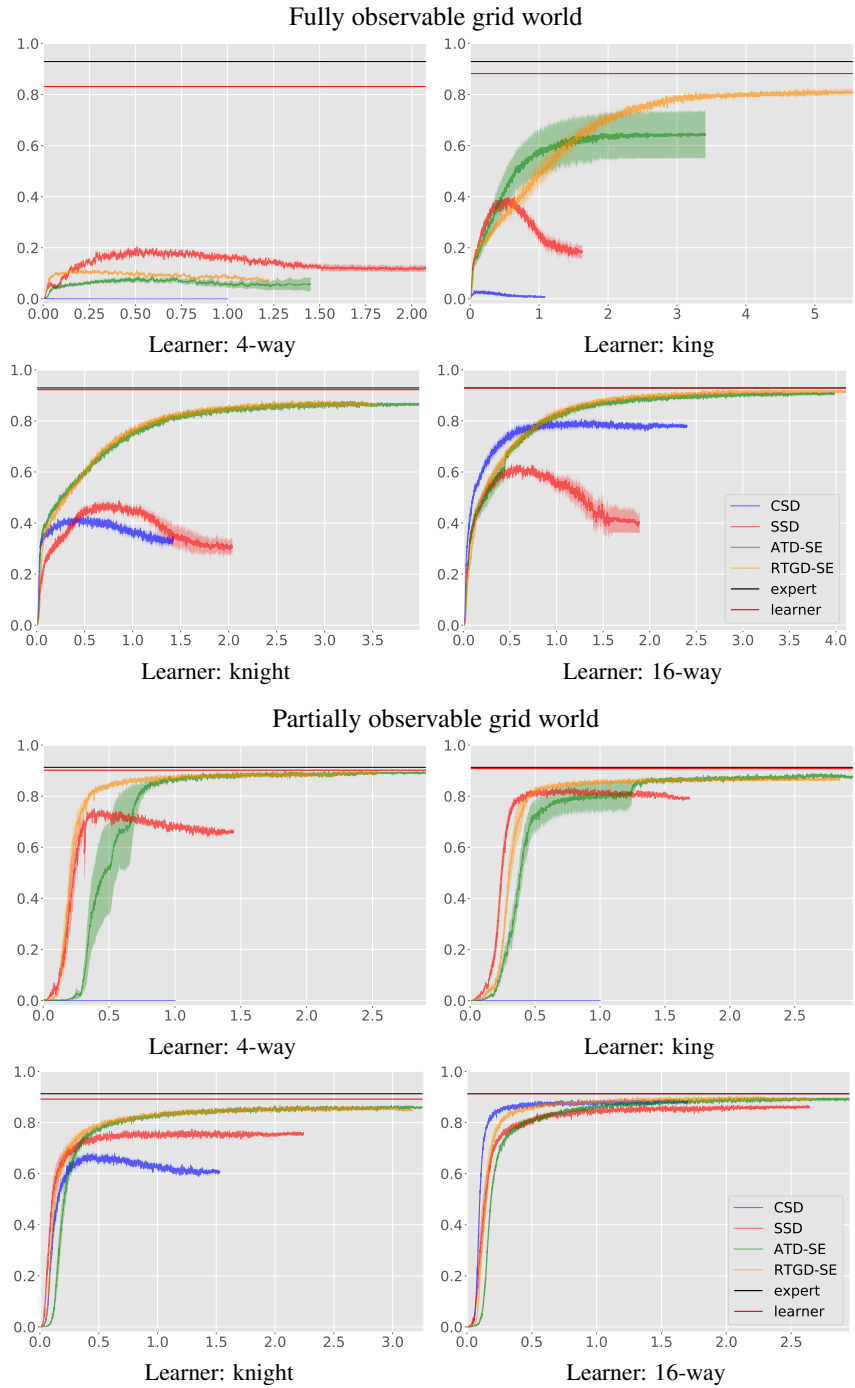
\centering
Fully observable grid world\\
\showplott{16-way}{mdp}
\vspace{0.3cm}\\
Partially observable grid world\\
\showplott{16-way}{pomdp}
\caption{Expert: 16-way}
\label{fig:appendix4}
\end{figure*}

\end{document}